\documentclass[lettersize,journal]{IEEEtran}
\usepackage[utf8]{inputenc}
\usepackage{amsmath,amsfonts}
\usepackage{bm}
\usepackage{algorithm,algorithmic}
\usepackage{array}
\usepackage[caption=false,font=normalsize,labelfont=sf,textfont=sf]{subfig}
\usepackage{textcomp}
\usepackage{stfloats}
\usepackage{url}
\usepackage{multirow}
\usepackage{array}  %
\usepackage{booktabs} 
\usepackage{etoolbox} 
\usepackage{makecell} %

\allowdisplaybreaks[4]

\usepackage{verbatim}
\usepackage{graphicx}
\def\BibTeX{{\rm B\kern-.05em{\sc i\kern-.025em b}\kern-.08em
    T\kern-.1667em\lower.7ex\hbox{E}\kern-.125emX}}
\usepackage{balance}

\usepackage{multirow} %
\usepackage[colorlinks,
            linkcolor=blue,
            anchorcolor=blue,
            citecolor=blue]{hyperref}%
\usepackage{amsmath, amssymb}%

\usepackage{xcolor}
\usepackage{soul}         %
\sethlcolor{yellow!20}
\soulregister\cite7
\soulregister\ref7

\usepackage[normalem]{ulem}
\definecolor{r1color}{HTML}{FFFFCC} %
\definecolor{r2color}{HTML}{C1FFC1} %
\definecolor{r3color}{HTML}{B2FFFF} %
\definecolor{r4color}{HTML}{FFB6C1} %
\definecolor{r5color}{HTML}{FFDAB9} %
\definecolor{r6color}{HTML}{E6E6FA} %

\hypersetup{hidelinks} %

\begin{document}

\title{A Dual-Domain Refinement Network with FBP-Based {Jacobian} Learning for Sparse-View Dual-Energy CT Material Decomposition}

\author{Qian Liu,  Xiaohong Fan, Ke Chen, Chong Chen, Shuaikang Wang, and Jianping Zhang
\thanks{This work was supported by the science and technology innovation Program of Hunan Province (2024RC9008) and Technology Department (2023GK2029), the National Natural Science Foundation
of China (No. 12501750, 
\protect{12322117 and 12288201), and the National Key Research \& Development Program of China through grant 2023YFA1009300.} Corresponding authors: \href{https://orcid.org/0000-0002-6093-6623}{K. Chen} and J. Zhang.}
\thanks{Q. Liu, S. K. Wang and J. Zhang are with the School of Mathematics and Computational Science, Xiangtan University, National Center for Applied Mathematics in Hunan, Key Laboratory for Intelligent Computing and Information Processing of the Ministry of Education, Xiangtan 411105, China (e-mail: \{liuqian,\;202431510150\}@smail.xtu.edu.cn, jpzhang@xtu.edu.cn).}
\thanks{X. H. Fan is with College of Mathematical Medicine, Zhejiang Normal University, Jinhua 321004, China (e-mail: fanxiaohong@zjnu.edu.cn).}
\thanks{K. Chen is with the Department of Mathematics and Statistics, University of Strathclyde, G1 1XH Glasgow, U.K. (e-mail: k.chen@strath.ac.uk).}
\thanks{C. Chen is with the \protect{State Key Laboratory of Mathematical Sciences, Academy of Mathematics and Systems Science, }Chinese Academy of Sciences, Beijing, 100190, China (e-mail: chench@lsec.cc.ac.cn).}
}
\markboth{Journal of \LaTeX\ Class Files,~Vol.~18, No.~9, September~2020}%
{How to Use the IEEEtran \LaTeX \ Templates}

\maketitle
\begin{abstract}
Dual-energy CT (DECT) exploits attenuation differences across different X-ray spectra to provide richer material information and has been widely used in medical imaging. While sparse-view acquisition can lower radiation exposure, it makes DECT material decomposition even more challenging, as the problem is nonlinear and ill-posed. Existing deep unrolling approaches generally do not explicitly incorporate the Jacobian operator induced by the nonlinear forward model, and their sparsity priors are still mainly built on conventional convolutions, which are insufficient for modeling global structural information.
This study addresses the challenge of DECT multi-material decomposition {in sparse-view settings} by representing it as a sparse-regularized nonlinear {least-squares} problem. To solve it, we propose an iterative dual-domain refinement network (DECT-DRNet). In each iteration, the {filtered back-projection (FBP)-based} Jacobian approximation module is {first used} to generate an intermediate material decomposition result. Here, we characterize {the forward model underlying} material decomposition using a nonlinear operator, and then construct a theoretically grounded learnable approximation of the adjoint Jacobian operator by integrating the FBP algorithm with a U-Net into the backward process. {In addition}, to address the {limited ability} of existing deep learning-based decomposition methods {to globally suppress} noise and artifacts, we introduce a learnable sparse dual-domain regularization term that incorporates Fourier {residual convolutional} blocks. This refinement block combines {geometric feature extraction} in the image domain with noise suppression in the frequency domain, allowing the model to capture both global and local features while maintaining structural details. In the sparse-view setting with 60 projection views, DECT-DRNet improves the {mean} PSNR by 3.23 dB and {2.01} dB over classical DECT material decomposition methods on the breast spectral CT and abdominal CT datasets, respectively, demonstrating its ability to achieve more accurate material decomposition under sparse-view conditions. The corresponding qualitative results further show that DECT-DRNet effectively suppresses noise and artifacts.
\end{abstract}

\begin{IEEEkeywords}
Sparse-view {dual-energy CT}, multi-material decomposition, iterative {unrolling} network, nonlinear inverse problem, FBP-based {Jacobian} learning, Fourier convolution.
\end{IEEEkeywords}

\section{Introduction}
\IEEEPARstart{C}{onventional} single-energy computed tomography (CT) frequently encounters difficulties in accurately {differentiating between materials} because it relies solely on the attenuation properties of X-rays {from a single energy spectrum.} In contrast, {dual-energy} CT (DECT)  acquires X-ray data using two different energy spectra, {thereby improving} {the ability to distinguish between materials}, and has demonstrated strong potential for clinical diagnosis and disease characterization \cite{Xu2018ImageDA,articleLazar,articleRamon,Muenzel2017MaterialDI}. Despite its excellent performance in medical imaging, DECT still faces a key {challenge}: the material decomposition task is inherently nonlinear \cite{chen2017image}, \cite{gao2024analysis}, \cite{gao2025convergence}. In clinical practice, sparse-view DECT is particularly attractive because it can substantially shorten scan times and reduce radiation dose; however, the severely ill-posed nature of this setting can degrade image quality \cite{zhang2024one}. Consequently, reducing patient radiation dose while preserving high-quality material decomposition in sparse-view scenarios has become a central research problem in DECT imaging.

Model-based iterative algorithms are currently the dominant approach for solving the DECT material decomposition problem. These methods generally formulate the task as an optimization problem that combines a data-fidelity term with prior regularization terms \cite{harms2016noise,wang2022iterative,fang2025virtual}, \cite{six2021gauss}. Commonly used priors include low-rank constraints \cite{lu2024projection}, sparsity \cite{ding2018image}, total variation (TV) \cite{chen2017image,chen2018algorithm}, and edge-preserving regularizations \cite{long2014multi,zhao2016using,xue2017statistical}. To solve the resulting optimization problems, classical numerical schemes such as the primal–dual (PD) method \cite{gao2022extended,chen2024prototyping}, the alternating direction method of multipliers (ADMM) \cite{ding2018image}, and the conjugate gradient (CG) method \cite{niu2014iterative} are widely used. These algorithms are supported by well-established {mathematical foundations} and {provide convergence guarantees}. Nevertheless, the complexity of the underlying models often requires a large number of iterations, which {leads} to considerable computational costs. In addition, the priors are usually {manually designed}, which can {limit their ability} to faithfully represent the {underlying characteristics} of DECT images, and tuning the associated hyperparameters remains a difficult and time-consuming task.

Recently, deep learning-based methods have attracted extensive attention for DECT material decomposition. Some methods exploit the strong {nonlinear} fitting capability inherent in deep learning to perform {this task} \cite{zhang2024one,su2022direct,Shi2020AMD}. For example, Clark et al. \cite{Clark2018MultienergyCD} developed a U-Net-based framework for DECT material decomposition. Gong et al. \cite{https://doi.org/10.1002/mp.14523} proposed Incept-Net, which employs multi-branch modules to {perform} three-material decomposition. Shi et al. \cite{shi2024multi} introduced a graph edge-conditioned convolutional network for DECT material decomposition that combines graph neural networks with convolutional neural networks, thereby enabling the joint extraction of local and nonlocal features. Wang et al. \cite{wang2025Clip} presented a CLIP-Driven generalizable model for DECT material decomposition. However, these approaches do not explicitly incorporate the intrinsic {nonlinearity} of the DECT imaging model into the network design. To better address this {limitation}, Zhang et al. \cite{https://doi.org/10.1002/mp.13489} proposed a dual-input and dual-output butterfly network that performs material decomposition based on the mapping between DECT images and {basis material images}. {Peng et al. \cite{https://doi.org/10.1002/mp.17255} proposed a generative adversarial network that incorporates a data-fidelity loss to perform image-domain material decomposition and reduce noise in DECT.} However, these methods, which are based on linear model assumptions, tend to amplify noise and introduce {artifacts}. This drawback is exacerbated in sparse-view acquisition, where the decomposition problem becomes more severely ill-posed, leading to {reduced} stability and accuracy {of} the material decomposition results. In addition, these methods are generally {limited to decomposing} two basis materials and {cannot be directly extended to decompose} three or more materials. {In recent years, diffusion models have demonstrated strong performance for dual-energy CT material decomposition \cite{jiang2025multi, vazia2026material}. However, these approaches typically entail high inference costs and may introduce undesired artifacts.}

The combination of deep learning with traditional optimization algorithms has achieved great success in image reconstruction \cite{adler2018learned,xiang2021fista,zhang2020amp,fan2023nest}. This paradigm fuses the interpretability and robustness of classical optimization with the computational efficiency and flexibility of deep learning. {Among such approaches, deep unrolling strategies incorporate iterative optimization schemes directly into network architectures. However, most of them have been {primarily designed} for linear or approximately linear forward models, and their use in nonlinear inverse problems such as DECT material decomposition remains relatively limited \cite{zhou2024deep}.} {Eguizabal et al.~\cite{eguizabal2022deep} introduced an unrolled iterative network for projection-domain material decomposition in spectral CT. Li et al.~\cite{https://doi.org/10.1002/mp.15817} developed an improved iterative neural network for DECT material decomposition.} Lantz et al. \cite{10230700} proposed a non-convex ADMM-based unrolled model for DECT {material} decomposition. Ma et al. \cite{ma2025dual}
 developed a physics-informed dual-domain learning framework that couples neural networks with an optimization algorithm for DECT material decomposition. {However, these strategies still exhibit two major limitations: (1)
One key challenge in DECT material decomposition is that the Jacobian operator of the nonlinear forward model varies with the current iterate, and its adjoint, which is needed for the gradient update, is computationally expensive to {compute} explicitly. Yet, this specific Jacobian operator has not been directly addressed in the existing literature. (2) Sparse regularization has not been sufficiently explored in existing methods. The few available learned regularization approaches mainly rely on conventional convolutions for representation learning, {which limits their ability to {suppress} global artifacts and {capture} large-scale structural information}~\cite{10376896,10.1007/978-3-031-43999-5_24}.}

To solve the above problems, we propose a novel interpretable iterative dual-domain refinement network based on a sparse prior (DECT-DRNet). {Specifically, we formulate} sparse-view DECT material decomposition as an {$\ell_1$}-regularized nonlinear least-squares problem and {unfold} the proximal gradient algorithm with a {filtered back-projection (FBP)-based} {Jacobian} approximation step into a learnable network architecture. {{From} the nonlinear forward model, we derive its Jacobian operator and the corresponding adjoint operator. Building on this derivation, we combine FBP with a pretrained U-Net to construct a theoretically grounded learnable {approximation to the} adjoint Jacobian operator.  In addition, we combine image-domain convolutions with Fourier residual convolutions to construct a dual-domain learned representation and impose a sparsity prior on this representation, enabling joint modeling of local structural information and global artifact patterns.}
The key contributions of our method are as follows.

\begin{enumerate}
\item {{We propose a novel DECT-DRNet} to solve the sparse-view DECT multi-material decomposition {problem}, {integrating} model-driven optimization and deep learning  {within a unified} iterative framework. It incorporates the nonlinear physical model, a gradient update step with a learnable {approximation of the adjoint Jacobian operator}, and learned sparse regularization.}
\item {We design an FBP-based Jacobian approximation module. To address the difficulty {in} directly computing the adjoint Jacobian operator {of} the nonlinear DECT forward model, we derive the Jacobian operator of the forward model and its corresponding adjoint {operator}. Based on this derivation, we combine {FBP} with a pretrained U-Net to construct a theoretically grounded learnable approximation of the adjoint Jacobian {operator} required {for} the gradient update.}
\item {We propose a dual-domain sparse regularization module built {on} a learnable geometric transformation to {implement the proximal mapping in} the proximal gradient algorithm. {In this module}, conventional convolutions are integrated with Fourier residual convolutions to simultaneously capture local structural details and globally correlated  patterns within a learned sparse representation.}
\end{enumerate}

\section{Nonlinear Inverse Problem for DECT Imaging}
DECT imaging acquires observed data {using} two distinct X-ray energy spectra, {{as} described by the following transmission equation}:
\begin{equation}\label{eq0a}
I_w(\ell)=I_0\int_E\:S_w(E)\exp\left(-\int_\ell\:\mu(\bm{s},E)\:d\zeta\right)\:dE,
\end{equation}
where $w=1,2$, $\ell$ is an X-ray path and {$d\zeta$ denotes the arc-length element along the path $\ell$.} The linear attenuation coefficient of the scanned object at point $\bm{s}=(s_1,s_2,s_3)$ and energy $E$ is denoted by $\mu(\bm{s},E)$. The $w$-th normalized effective spectrum $S_w(E)$ reflects the combined effect of the X-ray {tube} emission spectrum, the materials and thicknesses of the detector scintillator and filter, among other factors. The virtual monochromatic image at a given energy level $E$ is {related} to basis images through a material decomposition model described by
$$\mu(\bm{s},E)=\sum\limits_{n=1}^N\mu_n(E)\bm{x}_n(\bm{s}),$$
where $\bm{x}_n$ represents the volume fraction of the $n$-th material. {To simplify the nonlinear transmission model and express the measurements
in projection form, the measured X-ray intensities are transformed into the
post-log domain via the negative logarithm operation.} Consequently, the DECT reconstruction problem {can be formulated} as determining $\bm{x}_n(\bm{s})$ ($n=1,\dots,N$) from known polychromatic projections {given} by
\begin{equation*}
\begin{split}
\mathcal{A}_w(\ell)=&-\ln\frac{I_w(\ell)}{I_0}\\
=&-\ln\int_E S_w(E)\exp{\left(-\int_\ell\sum\limits_{n=1}^N\mu_n(E)\bm{x}_n\:d\zeta\right)}\:dE.
\end{split}
\end{equation*}

The discrete forward model, also referred to as the sampling model of DECT \cite{chen2024prototyping}, {\cite{gao2025afire}}, is expressed as follows: 
\begin{equation}\label{eq1}
  \mathcal{A}_{wi}(\boldsymbol{x})= -\ln\left(\sum_{m=1}^{M} S_{wm} \exp\left(-\sum_{j=1}^J a_{wij}\sum_{n=1}^{N} \mu_{mn}x_{nj}\right)\right),
\end{equation}
where $S_{wm}$ represents the normalized X-ray spectrum $w$ ($w=1, 2$) for the energy level $m$ ($m=1,2,...,M$); $\mu_{mn}$ is the linear attenuation coefficient {for material} $n$ ($n=1,2,...,N$) at energy $m$; $a_{wij}$ represents the path length crossed by ray $i$ ($i=1,2,...,I$) through voxel $j$ ($j=1,2,...,J$) within spectrum $w$; the vector $\boldsymbol{x} = (\boldsymbol{x}_1^T,\boldsymbol{x}_2^T,...,\boldsymbol{x}_J^T)^T\in\mathbb{R}^p$ ($p=J\times N$) with $T$ as {the} transpose operation, and $x_{nj}$ denotes the discrete value of the $n$-th basis material at the $j$-th pixel. 

Furthermore, we refer to $\boldsymbol{y}$ as observed projection data, which can alternatively be represented by the following nonlinear {multivariate} system:
\begin{equation}\label{eq2}
    \boldsymbol{y} =\mathcal{A}(\boldsymbol{x}),
\end{equation}
where $\mathcal{A} = ({\mathcal{A}_1}^T, {\mathcal{A}_2}^T)^{T}$ is a nonlinear operator that represents the forward model in (\ref{eq1}), and $\mathcal{A}_{w}=(\mathcal{A}_{wi})$ represents the vector of all X-ray paths $i$ under the energy spectrum $w$. However, the projection data typically contain noise or other components $\boldsymbol{\varepsilon}$, and {are  therefore} generally represented as
\begin{equation}\label{eq3}
    \boldsymbol{y} =\mathcal{A}(\boldsymbol{x})+\boldsymbol{\varepsilon}.
\end{equation}

Our focus is on solving the ill-posed inverse problem (\ref{eq3}), where we aim to determine $\bm{x}$ from the given projection data $\bm{y}$. This inverse problem {is more challenging than} the direct problem of {determining} $\bm{y}$ when $\bm{x}$ is known. Specifically, the DECT material decomposition problem requires solving for $\boldsymbol{x}$, which is a challenging nonlinear inverse problem. {In sparse-view settings, insufficient projection data significantly exacerbate the ill-posedness} of the problem, leading to severe artifacts in the material decomposition results. To tackle this challenge, regularization methods are generally employed to stabilize the solution process and {suppress} artifacts. 
Consequently, we formulate the problem as {an} $\ell_1$-regularized nonlinear least-squares problem as follows:
\begin{equation}\label{eq4}
\arg \min_{\boldsymbol{x}}\left\{f(\bm{x})+g_\tau(\bm{x}):=\frac{1}{2}\left\|\mathcal{A}(\boldsymbol{x})-\boldsymbol{y} \right\|_2^2+ g_\tau(\bm{x})\right\},
\end{equation}
where $g_\tau(\bm{x})=\tau \left\|\Psi \boldsymbol{x}\right\|_1$, $\tau$ is a regularization parameter{,} and $\Psi$ denotes the geometric transformation. {The regularization term enhances} the robustness of the solution by {imposing a} sparse prior {on} the transformation fields $\mathcal{T}=\{\boldsymbol{\phi}\in\mathbb{R}^q|\boldsymbol{\phi}=\Psi \boldsymbol{x}\}$.

The above regularization is often crucial for achieving stable solutions {to} inverse problems. However, determining the appropriate regularization parameter $\tau$ and the transformation $\Psi$ can be challenging and depends on the specific problem.

\section{Methodology}
In supervised learning, an algorithm {measures the error} through a loss function and seeks an appropriate function $\mathcal{F}$ to effectively reduce the error. This is often formulated as
\[\min_{\mathcal{F}}\mathbb{E}_{(\bm{x},\bm{y})\in\mathcal{D}}\left\|\boldsymbol{x}-\mathcal{F}(\boldsymbol{y})\right\|_2^2\text{ or }\min_{\mathcal{F}}\mathbb{E}_{\bm{x}\in\mathcal{X}}\left\|\boldsymbol{x}-\mathcal{F}(\mathcal{A}(\boldsymbol{x}))\right\|_2^2,
\]
where $\mathcal{A}(\cdot)$ is defined in {(\ref{eq2})}. The training dataset $\mathcal{D}=\mathcal{X}\times\mathcal{Y}$ consists of 
{observed data $\bm{y}\in\mathcal{Y}$ and the corresponding ground truth 
$\bm{x}\in\mathcal{X}$.} Deep unfolding networks are particularly useful in signal processing and image reconstruction when the optimization problem is well-structured and can be directly associated with a known algorithm. Generally, these networks are trained to learn a nonlinear optimization-structured mapping $h_{\boldsymbol{\theta}}(\cdot)$ using the learnable parameters $\boldsymbol{\theta}$, where $h_{\boldsymbol{\theta}}(\cdot)$ approximates the reconstruction mapping $\mathcal{F}(\cdot)$ by minimizing
\[\min_{\bm{\theta}}\mathbb{E}_{\bm{x}\in\mathcal{X}}\left\|\boldsymbol{x}-h_{\bm{\theta}}(\mathcal{A}(\boldsymbol{x}))\right\|_2^2.\]

To obtain {an optimization-driven} mapping {$h_{\boldsymbol{\theta}}(\cdot)$,} we begin by examining the reconstruction minimization problem (\ref{eq4}), which consists of a smooth data fidelity term and a non-smooth sparse regularization term. The proximal gradient method applied to (\ref{eq4}) is expressed as
\begin{equation}\label{eq4b}
\arg \min_{\boldsymbol{x}}\left\{\frac{1}{2}\left\|\bm{x}-(\boldsymbol{x}^{k-1}-\eta\nabla f(\bm{x}^{k-1}))\right\|_{{2}}^2+ {\eta} g_\tau(\bm{x})\right\}.
\end{equation}
If $\nabla f$ is Lipschitz continuous with a constant $L$, this approach has been shown to converge at a rate of $O(1/k)$ when using a fixed step length $\eta\in (0, 1/L]$ \cite{Combettes2011}. Consequently, {the problem can be solved} through a two-step iterative proximal method as
\begin{equation}\label{eq5}
\begin{split}
\boldsymbol{r}^k&=\varphi(\boldsymbol{x}^{k-1};\mathcal{A},\bm{y}, \eta)=\boldsymbol{x}^{k-1}-\eta\nabla f(\bm{x}^{k-1})\\&=\boldsymbol{x}^{k-1}-\eta {\mathrm{D}_{k-1}^{T}}(\mathcal{A}(\boldsymbol{x}^{k-1})-\boldsymbol{y})
\end{split}
\end{equation}
and
\begin{equation}\label{eq6}
\begin{split}
\boldsymbol{x}^k&=\psi(\bm{r}^k;\Psi,\lambda)=\arg\min_{\boldsymbol{x}}\frac{1}{2}\left\|\boldsymbol{x}-\boldsymbol{r}^k\right\|_2^2+\lambda\left\|\Psi \boldsymbol{x}\right\|_1,
\end{split}
\end{equation}
{where $\Psi$ is defined in (\ref{eq4}), $\lambda = \eta \tau$, and the Jacobian operator $\mathrm{D}_{k}$ of the nonlinear function $\mathcal{A}(\cdot)$ can be computed as}
\begin{equation}\label{eq7}
\mathrm{D}_k:=\mathrm{D}(\bm{x})|_{\bm{x}=\bm{x}^k} = \left.\left(\frac{ \partial \mathcal{A}_{wi}(\bm{x}) }{ \partial x_{nj}}\right)\right|_{\bm{x}=\bm{x}^k},
\end{equation}
where
\begin{align*}
\dfrac{ \partial \mathcal{A}_{wi} }{ \partial x_{nj}}\!&=\!\frac{\sum\limits_{m=1}^{M}\mu_{mn}a_{wij}S_{wm}\exp{\left(-\sum\limits_{j=1}^{J}a_{wij}\sum\limits_{n=1}^{N}\mu_{mn}x_{nj} \right)}}{\sum\limits_{m=1}^{M}S_{wm}\exp{\left(-\sum\limits_{j=1}^{J}a_{wij}\sum\limits_{n=1}^{N}\mu_{mn}x_{nj} \right)}}\\
\!&=\sum\limits_{m=1}^{M}\mu_{mn}a_{wij}S_{wm}+\!\sum\limits_{m=1}^{M}\mu_{mn}a_{wij}S_{wm}\epsilon_{wmi},\\
\end{align*}
and $\epsilon_{wmi}=\tau_{wmi}-1$ with
$$\tau_{wmi}=\cfrac{\exp{\left(-\sum\limits_{j=1}^{J}a_{wij}\sum\limits_{n=1}^{N}\mu_{mn}x_{nj} \right)}}{\sum\limits_{m=1}^{M}S_{wm}\exp{\left(-\sum\limits_{j=1}^{J}a_{wij}\sum\limits_{n=1}^{N}\mu_{mn}x_{nj}\right)}}.$$
A fixed-point iterative scheme can be developed as 
\begin{equation}
\boldsymbol{x}^{k}=\mathcal{H}_{\bm{\theta}}(\boldsymbol{x}^{k-1}):=\psi(\varphi(\boldsymbol{x}^{k-1};\mathcal{A},\bm{y}, \eta);\Psi,\lambda),
\label{eq-fixed}
\end{equation}
where {$\boldsymbol{\theta}=\{\Psi,\eta,\lambda\}$ denotes the set of parameters, and} $\bm{y}=\mathcal{A}(\bm{x}^*)$.
If $\mathcal{H}_{\bm{\theta}}(\cdot)$ is a contraction mapping, then there exists a solution $\boldsymbol{x}^*$ that satisfies
\[\boldsymbol{x}^{*}=\mathcal{H}_{\bm{\theta}}(\boldsymbol{x}^*)
  =h_{\boldsymbol{\theta}}(\boldsymbol{y})=h_{\boldsymbol{\theta}}(\mathcal{A}(\boldsymbol{x}^*))\approx \mathcal{F}(\mathcal{A}(\boldsymbol{x}^*)).\]

The above model-based optimization techniques use iterative methods to progressively refine solutions, thus increasing accuracy with each step $k$. Although these methods are effective for tackling problems that cannot be solved by analytical techniques alone and provide good interpretability and flexibility, they also involve significant computational demands and require careful selection of algorithm parameters.
{We then model the} step {in} (\ref{eq-fixed}) by integrating traditional optimization techniques with residual learning.

The effectiveness of {the} above optimization methods is largely dependent on the appropriate selection of both regularization and algorithm parameters, as exemplified by $\ell_1$-regularization (\ref{eq4}) and the iterative optimization method (\ref{eq5})-(\ref{eq6}). Moreover, when using an iterative technique to solve the proximal-point subproblem, {proper} tuning of the regularization parameter $\lambda$ (or $\tau$) and the algorithm parameter $\eta$ is essential. {In particular,} in the proximal gradient iterative regularization method involving the nonlinear system (\ref{eq2}), typical parameter selection strategies might not yield satisfactory results. {Furthermore,} utilizing a sparse prior for the transformation fields $\mathcal{T}=\{\boldsymbol{\phi}\in\mathbb{R}^q|\boldsymbol{\phi}=\Psi \boldsymbol{x}\}$ can help prevent overfitting by focusing on the most important features, {thereby improving} generalization {to} {unseen} data. Therefore, {in image processing, selecting a suitable sparse transformation $\Psi$ is crucial} for creating efficient, interpretable, and robust representations that can enhance various computational and analytical tasks.

In this study, we propose a deep unfolding network inspired by the above iterative algorithm to tackle nonlinear {DECT} basis-material decomposition problems, which involves ``unfolding" the fixed-point algorithm (\ref{eq-fixed}) into a finite series of layers, where each layer represents one iteration of the algorithm. This strategy trains the network to learn the parameters $\boldsymbol{\theta}=\{\Psi,\eta,\lambda\}$ of the unfolded algorithm, rather than keeping them fixed or tuning them manually, so that
\[
\mathcal{E}(\mathcal{X},\mathcal{A})=\inf_{\boldsymbol{\theta}} \mathbb{E}_{\bm{x}\in\mathcal{X}}\left\{\mathcal{E}_0(\boldsymbol{x}; \boldsymbol{\theta}, \mathcal{A}):=\left\|\boldsymbol{x}-h_{\boldsymbol{\theta}}(\mathcal{A}(\boldsymbol{x}))\right\|_2^2\right\}. 
  \]
In other words, the parameters are optimized by minimizing the expected reconstruction error {over} the distribution $\mathcal{X}$.
  
The proposed architecture combines the mathematical foundation of traditional iterative algorithms with the adaptive learning and flexibility of deep neural networks. After training, these networks  typically require fewer iterations or layers to obtain high-quality solutions, {thereby achieving} faster convergence compared to traditional iterative methods and potentially {improving} performance by learning from data.

\subsection{FBP-Based Jacobian Learning Module} 
In DECT imaging, $\mathcal{A}(\cdot)$ {denotes} the nonlinear forward process, {which can be readily evaluated.} {Applying the weighted Jensen's inequality to the discrete forward model in (\ref{eq1}) yields} the following linear upper bound
\begin{align*}\label{eq1d}
\mathcal{A}_{wi}(\boldsymbol{x})=& -\ln\left(\sum_{m=1}^{M} S_{wm} \exp\left(-\sum_{j=1}^J a_{wij}\sum_{n=1}^{N} \mu_{mn}x_{nj}\right)\right)\\
\leq&-\sum_{m=1}^{M} S_{wm} \ln\left(\exp\left(-\sum_{n=1}^{N} \mu_{mn}\sum_{j=1}^J a_{wij}x_{nj}\right)\right)\\
=&\sum_{j=1}^J a_{wij}\left(\sum_{m=1}^{M} S_{wm} \left(\sum_{n=1}^{N} \mu_{mn}x_{nj}\right)\right){=[\mathcal{P}\mathcal{L}(\bm{x})]_{wi}},
\end{align*} 
{where the spectral fusion operator $\mathcal{L}(\cdot)$ and the projection operator $\mathcal{P}(\cdot)$ are defined by}
\begin{equation*}
\begin{split}
\mathcal{L}(\bm{x}) 
 &= ({\mathcal{L}_{1}}(\bm{x}_1),\dots,{\mathcal{L}_{1}}(\bm{x}_J), {\mathcal{L}_{2}}(\bm{x}_1),\dots,{\mathcal{L}_{2}}(\bm{x}_J))^{T}\in\mathbb{R}^{2J},\\
\mathcal{P}(\bm{\xi})
&=(\bm{a}^J_{11}\cdot\bm{\xi}^1,\dots,\bm{a}^J_{1I}\cdot\bm{\xi}^1,\bm{a}^J_{21}\cdot\bm{\xi}^2,\dots,\bm{a}^J_{2I}\cdot\bm{\xi}^2)^T\in\mathbb{R}^{2I}\\
\end{split}    
\end{equation*}
{and $\boldsymbol{\mu}_m=(\mu_{m1},\mu_{m2},...,\mu_{mN})\in\mathbb{R}^N$,}
\begin{align*}
&\mathcal{L}_{w}(\bm{x}_j)= \sum_{m=1}^{M} S_{wm}\boldsymbol{\mu }_m^T\bm{x}_j{,}\;\bm{a}^J_{wi}=({a}_{wi1},\dots,{a}_{wiJ})^T\in\mathbb{R}^J,\\
&\boldsymbol{\xi}^w=(\xi^w_{1},\xi^w_{2},...,\xi^w_{J})=({\mathcal{L}_{w}}(\bm{x}_1),\dots,{\mathcal{L}_{w}}(\bm{x}_J))\in\mathbb{R}^J.
\end{align*}
Therefore, {the nonlinear forward operator $\mathcal{A}(\cdot)$ in (\ref{eq3})} can be decomposed {as follows}:
\begin{equation}\label{eq8}
\mathcal{A}(\cdot)=\mathcal{P}\mathcal{L}(\cdot)+{\delta (\cdot)}, 
\end{equation}
{where
${\delta(\cdot)}=\mathcal{A}(\cdot)-\mathcal{P}\mathcal{L}(\cdot)$ is defined as the nonlinear residual term, which characterizes the gap between the nonlinear forward operator and the Jensen-based linear surrogate $\mathcal{P}\mathcal{L}(\cdot)$. This decomposition is adopted as a heuristic modeling strategy.} However, the forward process $\mathcal{A}(\bm{x})$ is still calculated using the nonlinear system (\ref{eq1}). The purpose of decomposing $\mathcal{A}(\cdot)$ in (\ref{eq8}) is to facilitate the derivation of an approximation {to} the adjoint Jacobian operator ${\mathrm{D}_k^T(\cdot)}$ associated with $\mathcal{A}(\cdot)$.

According to (\ref{eq5}), generating the intermediate material decomposition result $\boldsymbol{r}^k$ {typically} involves {the} nonlinear operator $\mathcal{A}(\cdot)$ and its adjoint Jacobian operator ${\mathrm{D}_k^T(\cdot)}$. However, {this operator} ${\mathrm{D}_k^T(\cdot)}$ is based on the result of the preceding iteration $\boldsymbol{x}^{k}$, and it is more difficult to {compute} directly. We {decompose} the adjoint Jacobian operator ${\mathrm{D}_k^T(\cdot)}$ of $\mathcal{A} (\cdot)$ {as follows: } 
 \begin{align}\label{eq9a}
{\mathrm{D}_k^T(\boldsymbol{z})} = {(\nabla\mathcal{A}(\boldsymbol{x}^k))^{T}(\boldsymbol{z})}\nonumber&=\mathcal{L}^
{T}\mathcal{P}^{T}(\boldsymbol{z}) + {(\nabla \delta\left(\bm{x}^{k}\right))^T(\boldsymbol{z})},
\end{align}
where
\begin{equation}
{(\nabla \delta\left(\bm{x}^{k}\right))^T} =\left(\left[\sum\limits_{m=1}^{M}\mu_{mn}a_{wij}S_{wm}{\epsilon_{wmi}^k}\right]_{nj,wi}\right)\in\mathbb{R}^{p\times 2I},
\label{eq9a2}
\end{equation}
{with $\epsilon_{wmi}^{k}$ obtained by evaluating $\epsilon_{wmi}$ in (\ref{eq7}) at the current iterate $\bm{x}^{k}$.}

The linear {backprojection} is defined by
\[(\mathcal{L}^
{T}\mathcal{P}^{T})(\bm{z})=((\mathcal{L}^
{T}\mathcal{P}^{T})_{1\cdot}(\bm{z});\dots;(\mathcal{L}^
{T}\mathcal{P}^{T})_{J\cdot}(\bm{z}))\in\mathbb{R}^p,\]
and the vector of the basis materials at voxel $j$ is given by
\[(\mathcal{L}^
{T}\mathcal{P}^{T})_{j\cdot}(\bm{z})=\sum\limits_{m=1}^M\sum\limits_{w=1}^2S_{wm}(\bm{a}_{wj}^I\cdot\bm{z}^w)\bm{\mu}_m^T\in \mathbb{R}^N,\]
{where} the backprojection operator $\mathcal{P}^{T}(\cdot)$ and the basis material decomposition operator $\mathcal{L}^{T}(\cdot)$ are defined by
\begin{equation*}
\begin{split}
\mathcal{P}^{T}(\bm{z})
&=(\bm{a}^I_{11}\cdot\bm{z}^1,\dots,\bm{a}^I_{1J}\cdot\bm{z}^1,\bm{a}^I_{21}\cdot\bm{z}^2,\dots,\bm{a}^I_{2J}\cdot\bm{z}^2)^T,\\
\mathcal{L}^T(\bm{\eta}) 
&=\left(\sum_{m=1}^{M} \sum\limits_{w=1}^2S_{wm}\eta^1_w\boldsymbol{\mu }_m^T,\dots,\sum_{m=1}^{M} \sum\limits_{w=1}^2S_{wm}\eta^J_w\boldsymbol{\mu }_m^T\right)^T,
\end{split}    
\end{equation*}
{with} $\mathcal{P}^{T}(\bm{z})\in\mathbb{R}^{2J}$, $\mathcal{L}^T(\bm{\eta})\in\mathbb{R}^{p}$, and
\begin{equation*}
\begin{split}
&\bm{z}=(\bm{z}^1;\bm{z}^2)\in \mathbb{R}^{2I},\;\;
\bm{a}^I_{wj}=({a}_{w1j},\dots,{a}_{wIj})^T\in\mathbb{R}^I,\\
&\bm{\eta}=(\bm{\eta}^1;\dots;\bm{\eta}^J),\;\;\bm{\eta}^j=(\eta^j_1;\eta^j_2)^T,\;\;\eta^j_w=\bm{a}^I_{wj}\cdot\bm{z}^w.
\end{split}    
\end{equation*}

{Furthermore, we propose to approximate 
${(\nabla \delta\left(\bm{x}^{k}\right))^T(\boldsymbol{z})}$ as}
$${(\nabla \delta\left(\bm{x}^{k}\right))^T(\boldsymbol{z})}\approx \bm{\Delta}_k\circ\mathcal{L}^T \mathcal{P}^{T}(\bm{z}),$$
{using a learnable mapping ${\bm{\Delta}_k}$. This approximation is designed to 
generalize the Jacobian of the residual term while incorporating the 
measurement physics through the operator $\mathcal{L}^{T}\mathcal{P}^{T}$.}

{Accordingly, the adjoint Jacobian operator ${\mathrm{D}_k^T(\cdot)}$ of $\mathcal{A}(\cdot)$ can be written as} 
\begin{equation*}\label{eq9}
{{\mathrm{D}_k^T(\boldsymbol{z})}}=\mathcal{L}^
{T}\mathcal{P}^{T}(\bm{z}) + {(\nabla \delta\left(\bm{x}^{k}\right))^T(\boldsymbol{z})}
\approx (\mathcal{I}+\bm{\Delta}_k)\circ\mathcal{L}^T\mathcal{P}^{T}(\bm{z}).
\end{equation*}

{To improve convergence efficiency and reconstruction quality under a limited number of optimization iterations, iterative CT reconstruction methods frequently employ analytic reconstruction operators as approximate implementations of the adjoint of the projection operator \cite{gao2016fused}. Typical examples {include the} FDK operator for CBCT, the Katsevich exact reconstruction operator for helical CT, and the FBP operator for two-dimensional fan-beam CT \cite{chen2020airnet}. Because $\mathcal{P}^{T}(\cdot)$ represents the adjoint of the CT projection operator and our study focuses on a two-dimensional fan-beam scanning configuration, we also replace it with the FBP operator in this work, {in which} the ramp filter is employed as a key component. }

Taking advantage of the strong nonlinear approximation capability of convolutional neural networks, we pretrain a U-Net to learn the mapping $(\mathcal{I}+\bm{\Delta}_k)\circ\mathcal{L}^T(\cdot)$. Following the U-Net design described in \cite{10.1007/978-3-319-24574-4_28}, we incorporate batch normalization (BN) layers~\cite{Ioffe2015BatchNA} between the convolutional layers to accelerate the training process.

Consequently, we combine FBP with the pretrained U-Net to approximate the  {adjoint Jacobian} operator {${\mathrm{D}_k^T(\cdot)}$} via
\begin{equation}\label{eq10}
{{\mathrm{D}_k^T(\boldsymbol{z})}}\approx \text{U-Net} \circ \text{FBP}(\bm{z}).
\end{equation}
Note that reference~\cite{10230700} also adopts a hybrid FBP–U-Net framework, but in that work, it is used solely to provide an initial estimate. In contrast, our method applies this architecture to approximate the  {adjoint Jacobian} operator  {${\mathrm{D}_k^T(\cdot)}$} within each iteration of the proximal gradient scheme. In addition, the parameter $\eta$ is treated as a learnable quantity. Finally, the FBP-based  {Jacobian} learning module is depicted in Fig.~\ref{fig1}(a).

\subsection{Dual-Domain Geometric Refinement Module}
{Effectively solving} the proximal mapping problem is crucial. Traditionally, the optimization problem (\ref{eq6}) involves {manually designing} the sparse geometric transformation $\Psi$ and {selecting} the parameter $\lambda$, which can be {time-consuming and challenging.}  {Inspired by {ISTA-Net}~\cite{8578294}, we parameterize $\Psi$ as a learnable nonlinear mapping {$\mathcal{G}$ and} make the regularization parameter $\lambda$  learnable.}
Accordingly, (\ref{eq6}) can be rewritten for the $k$-th iteration as
  \begin{equation}\label{eq11}
\boldsymbol{x}^k=\arg\min_{\boldsymbol{x}}\frac{1}{2}\left\|\boldsymbol{x}-\boldsymbol{r}^{k}\right\|_2^2+\lambda^k\left\|\mathcal{G}(\boldsymbol{x})\right\|_1.
\end{equation}
Then, {based on} the ISTA-Net theory~\cite{8578294}, {(\ref{eq11}) can be approximated in the learned transform domain as follows{:}}  
\begin{equation}\label{eq12}
\boldsymbol{x}^k=\arg\min_{\boldsymbol{x}}\frac{1}{2}\left\|\mathcal{G}(\boldsymbol{x})-\mathcal{G}(\boldsymbol{r}^{k})\right\|_2^2+\rho^k\left\|\mathcal{G}(\boldsymbol{x})\right\|_1,
\end{equation}
where $\rho^k$ is the learnable threshold related to $\lambda^k$.

Finally, the solution to this problem (\ref{eq12}) (or (\ref{eq6})) can be derived as follows{:}
\begin{equation}\label{eq13}
\boldsymbol{x}^{k}=\tilde{\mathcal{G}}({\mathrm{soft}}(\mathcal{G}(\boldsymbol{r}^k),\rho^k)),
\end{equation}
{where the decoder $\tilde{\mathcal{G}}$ is the left inverse of the encoder $\mathcal{G}$, and $\mathrm{soft}(\cdot,\rho^k)$ denotes the soft-thresholding operator with threshold $\rho^k$, corresponding to the shrinkage step induced by {an $\ell_1$ sparsity prior} in the learned transform domain $\mathcal{G}$. 

We now describe the motivation and architecture of $\mathcal{G}$. {In} sparse-view DECT, the intermediate material images produced by the FBP-based Jacobian learning module still exhibit structured artifacts that are globally distributed. Because conventional image-domain convolutions primarily depend on local receptive fields to capture neighborhood correlations, they are inherently limited in their ability to represent and suppress such globally distributed artifacts~\cite{10376896}. Prior work has demonstrated that performing convolution in the Fourier domain is effective for removing global artifacts~\cite{10.1007/978-3-031-43999-5_24}. Inspired by this, we design $\mathcal{G}$ as a learnable nonlinear mapping that integrates image-domain convolutions with Fourier residual convolutional blocks, thereby jointly modeling local structural details in the spatial domain and global correlation patterns in the Fourier domain. A sparsity prior is then enforced on the resulting dual-domain representation {$\mathcal{G}(\bm{x})$}. Consequently, Eq.~(\ref{eq13}) provides a learned, parameterized form of the proximal mapping, where the transform $\mathcal{G}$, the soft-thresholding operation, and the synthesis transform $\tilde{\mathcal{G}}$ work together to implement the dual-domain geometric refinement module.}

\begin{figure*}[!ht]
\centering
\includegraphics[width=0.95\linewidth]{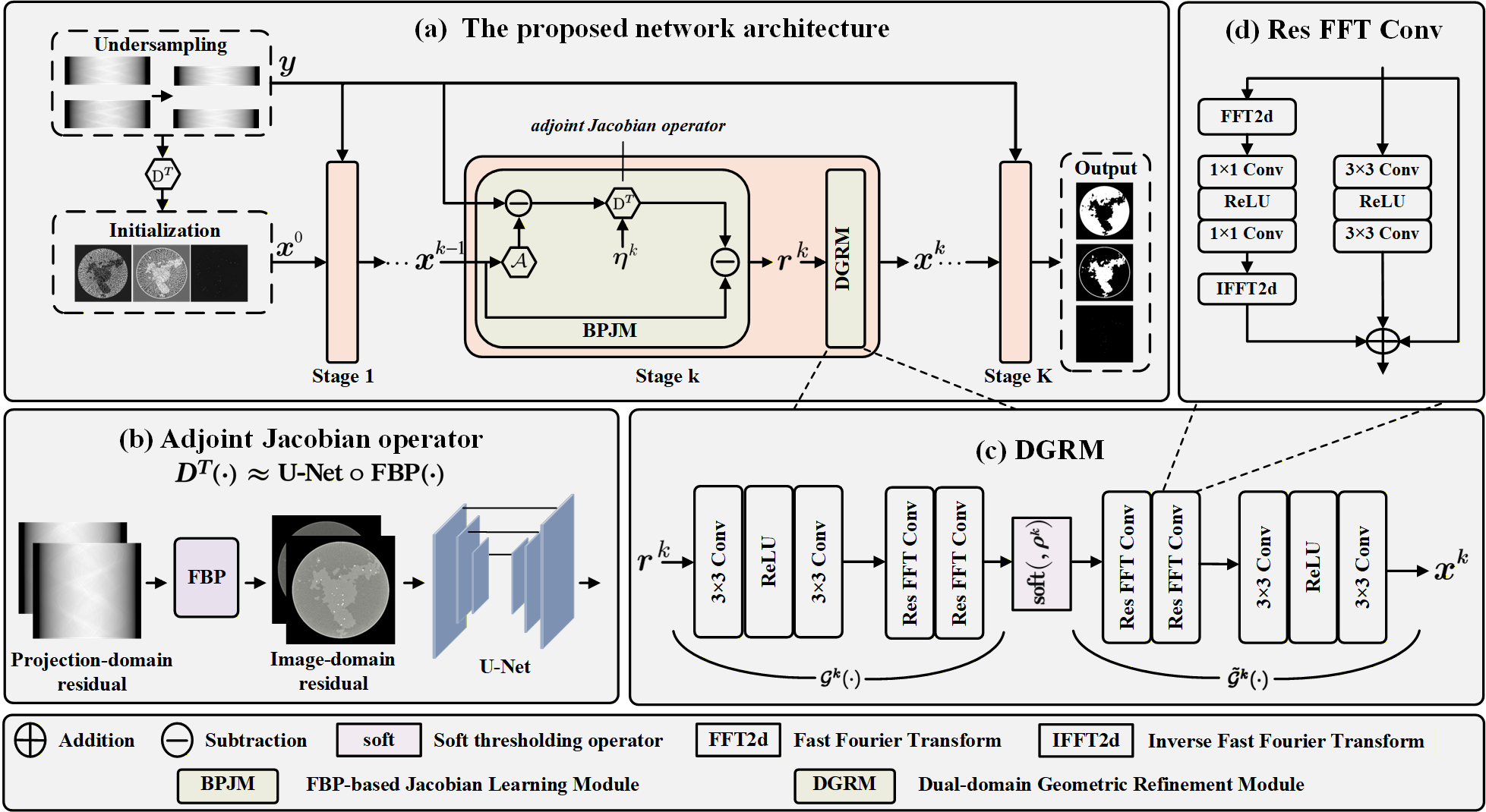} 
\caption{The proposed iterative dual-domain refinement network (DECT-DRNet). (a) The proposed network architecture; (b) The adjoint Jacobian operator $\mathrm{D}^T(\cdot)$ in FBP-based Jacobian learning module (BPJM); (c) The dual-domain geometric refinement module (DGRM); (d) The res FFT block.} \label{fig1}
\end{figure*}

\begin{algorithm}[!ht]
\caption{The Proposed DECT-DRNet Framework $\mathcal{N}$.}
\label{alg:DECT-DRNet}
\begin{algorithmic}
\STATE \underline{\textbf{Input}}: the physical parameters $\{S_{wm},\mu_{mn}\}_{w,m,n=1}^{2,M,N}$, the 
projection operator $\mathcal{P}$, the stage number $K$, {the} FBP 
operator, {the} pretrained U-Net, {and} the training dataset
$\mathcal{D} = \left\{ \left( \boldsymbol{x}_\ell, \boldsymbol{y}_\ell\right) \right\}_{\ell=1}^{L}$.

\STATE \underline{\textbf{Initialize}}: $\boldsymbol{x}^0 = \text{U-Net} \circ \text{FBP}(\boldsymbol{y})$, {and} the learnable parameters\\
$\Theta = \{\rho^k, \eta^k, {\mathcal{G}^k, \tilde{\mathcal{G}}^k}\}_{k=1}^{K}$. 

\STATE \underline{\textbf{Inference}}:

\STATE 1: \textbf{for} $k = 1, 2, \ldots, K$ \textbf{do}
\STATE 2:\qquad$\boldsymbol{r}^k=\boldsymbol{x}^{k-1}-\eta^k
(\text{U-Net} \circ \text{FBP}(\mathcal{A}(\boldsymbol{x}^{k-1})-\boldsymbol{y})),$ 
\STATE 3:\qquad$\boldsymbol{x}^k={\tilde{\mathcal{G}}^k}({\mathrm{soft}}({\mathcal{G}^k}(\boldsymbol{r}^k),\rho^k)). $ 
\STATE 4: \textbf{end for}

\STATE \underline{\textbf{Training}}:
\STATE 1:\qquad$\mathcal{L}_{{\text{total}}} = \mathcal{L}_{{\text{discrepancy}}}+\gamma \mathcal{L}_{{\text{constraint}}}$;
\STATE \underline{\textbf{Output}}: $\boldsymbol{x}^K=\mathcal{N}(\bm{y};\Theta)$.
\end{algorithmic}
\end{algorithm}

The learnable nonlinear geometric refinement mapping $\mathcal{G}$ is constructed as follows. First, $\bm{r}^{k}$ is passed through two CNN layers with a ReLU activation between them to suppress noise and extract local texture features. However, due to the inherently local receptive fields of CNNs, this alone is insufficient to model global information. To address this, we integrate the Fourier residual convolutional block \cite{Mao2021IntriguingFO}, illustrated in Fig. \ref{fig1}(d). This dual-domain feature extraction block introduces a residual frequency branch that converts feature maps into the frequency domain for learning, thereby improving the network’s ability to encode global information. The resulting feature maps are then processed by two {Fourier residual convolutional blocks.} The left inverse $\tilde{\mathcal{G}}$ is designed as a symmetric counterpart of $\mathcal{G}$, comprising two Fourier residual convolutional blocks followed by two CNN layers separated by a ReLU activation. 

{Moreover, all unrolling stages employ an identical architectural design, while each stage is equipped with its own learnable transform with independent parameters. This enables stage-dependent transformations that can adapt to the characteristics of each step in the iterative update. In particular, the $k$-th stage is associated with a learnable transform $\mathcal{G}^k$. Thus, the output at the $k$-th unrolling stage is defined as}
\begin{equation}\label{eq14}
\boldsymbol{x}^{k}=\tilde{\mathcal{G}}^k({\mathrm{soft}}(\mathcal{G}^k(\boldsymbol{r}^k),\rho^k)).
\end{equation}
{The overall dual-domain geometric refinement module, which models the proximal-point subproblem}, is depicted in Fig. \ref{fig1}(c). 

The overall architecture of DECT-DRNet is summarized in Algorithm~\ref{alg:DECT-DRNet} and illustrated in Fig.~\ref{fig1}. This network is designed to {perform} DECT material decomposition using an interpretable Fourier-based iterative unrolling strategy {that combines} FBP-based {Jacobian} learning in the pre-reconstruction step {with} sparse regularization in the geometric transformation field $\mathcal{T}_\mathcal{G}=\{\boldsymbol{\phi}\in\mathbb{R}^q|\boldsymbol{\phi}=\mathcal{G}(\boldsymbol{x})\}$.

\subsection{Loss Function}
The total loss of the proposed framework {follows that used in} ISTA-Net~\cite{8578294}. It consists of a discrepancy loss and a constraint loss. Accordingly, the total loss is defined as follows{:}
\begin{align*}
\mathcal{L}_{{\text{total}}} &= \mathcal{L}_{{\text{discrepancy}}}+\gamma \mathcal{L}_{{\text{constraint}}},
\end{align*}
where
\begin{align*}
        \mathcal{L}_{{\text{discrepancy}}} &=\frac{1}{N{J}L}\sum_{\ell=1}^{L}\left\|\hat{\boldsymbol{x}}_{\ell}^{K}-\boldsymbol{x}_{\ell}\right\|_2^2, \\
        \mathcal{L}_{{\text{constraint}}}&=\frac{1}{N{J}L}\sum_{\ell=1}^{L}\sum_{k=1}^{K}\left\|\widetilde{\mathcal{G}}^{k}(\mathcal{G}^{k}(\boldsymbol{x}_{\ell}))-\boldsymbol{x}_{\ell}\right\|_2^2, 
\end{align*}
and $\hat{\boldsymbol{x}}_{\ell}^{K}$ denotes the reconstruction of the $\ell$-th image after the $K$-th iteration ($\ell=1,2,\dots,L$), with $\boldsymbol{x}_{\ell}$ denoting its ground truth. $N$ is the number of basis material components, {$J$} is the number of pixels per image, and $\gamma$ is a weighting factor that balances the relative contribution of the two loss terms.

\section{Experiments}
\subsection{Experimental Setup}
\subsubsection{Dataset}
{We conduct numerical experiments on two datasets to evaluate the performance of the proposed DECT-DRNet. The first is a breast spectral CT dataset obtained from the 2022 AAPM Deep Learning Spectral CT Grand Challenge~\cite{sidky2024report}. The second is an abdominal dataset, derived from the Low-Dose CT Challenge~\cite{mccollough2017low}.

{For the breast spectral CT dataset, we use simulated data} rather than patient-level clinical CT data. {Therefore, a patient count is not applicable to this dataset.} Each sample includes three tissue maps: {adipose}, {fibroglandular}, and {calcification}. The images have a resolution of $512 \times 512$ pixels, with an in-plane pixel spacing of approximately $0.3516\,\mathrm{mm}$ in both width and height, and all pixel intensities are normalized to the range $[0,1]$. The dataset contains $1000$ samples, which are randomly split into $800$ for training, $100$ for validation, and $100$ for testing. For the abdominal dataset, we select $1100$ abdominal CT slices of size $512 \times 512$ from $9$ patients. To obtain the reference material images, each slice is segmented into water and bone material images using a threshold-based approach \cite{xu2025direct,krahenbuhl2011efficient}, followed by intensity normalization. The physical pixel spacing and the pixel value range of these material images are matched to those of the breast spectral CT dataset. This dataset is partitioned {at the} patient level, with 849 slices from $7$ patients used for training, $93$ slices from $1$ patient for validation, and $158$ slices from $1$ patient for testing.

The mass attenuation coefficients for both datasets are taken from the X-ray mass attenuation coefficient database maintained by the National Institute of Standards and Technology (NIST)~\cite{hubbell1995tables}. The X-ray spectra for the breast spectral CT dataset are generated {using} SpekPy~\cite{bujila2020validation}, while the spectra for the abdominal dataset are produced using SpectrumGUI (available at https://sourceforge.net/projects/spectrumgui/). For the breast spectral CT dataset, the low- and high-kVp settings are 50 kVp and 80 kVp, respectively. For the abdominal dataset, the corresponding tube {voltage settings} are 80 kVp (low-kVp) and 140 kVp (high-kVp). Forward projections are generated using the projection tool from the AAPM Spectral CT Challenge~\cite{sidky2024report}. A two-dimensional fan-beam geometry is employed, with rapid kVp {switching} between low- and high-kVp views. The source-to-detector distance is 100 cm, and the source-to-rotation-center distance is 50 cm. The detector comprises 1024 elements, and its overall length is chosen according to the specific dataset. For the breast spectral CT dataset, the detector length is {determined from} the largest inscribed circle within the image region, {resulting in} a detector element size of approximately $0.358\,\mathrm{mm}$. For the abdominal dataset,  the detector length {is determined from} the smallest {circle circumscribing} the image region, , {resulting in} a detector element size of approximately $0.514\,\mathrm{mm}$.}

{To evaluate the reconstruction performance of DECT-DRNet under different sparse-view conditions, we employ four sparse-view configurations, {in which} the number of projection views per energy level is set to 15, 30, 60, or 90. For each sparse-view setting, both low-kVp and high-kVp sinograms are uniformly acquired over a full $360^\circ$ scanning range. The high-kVp projection angles are shifted by half of the angular sampling interval relative to the low-kVp projection angles, thereby simulating an acquisition scheme where the tube voltage alternates between adjacent projection views. To evaluate the robustness of DECT-DRNet under noisy projections, zero-mean Gaussian noise is added to the simulated projection data in the noise-robustness experiments. The detailed noise model and corresponding noise levels are described in the associated experimental section.}

\subsubsection{Training Details}
{
To assess the performance of the proposed method, we first evaluate DECT-DRNet on the breast spectral CT dataset and compare it with the traditional DECT material decomposition approach (Matrix Inversion) as well as five classical deep learning-based DECT material decomposition methods: Clark-UNet~\cite{Clark2018MultienergyCD}, DnCNN~\cite{lu2019learning}, FC-DenseNet~\cite{wu2019multi}, Incept-Net~\cite{https://doi.org/10.1002/mp.14523}, and GECCU-Net~\cite{shi2024multi}. The experiments on the breast spectral CT dataset are conducted under sparse-view settings with 15, 30, 60, and 90 projection views. For the abdominal dataset, we additionally compare the proposed method with the same five deep learning-based DECT material decomposition methods and a model-driven primal-dual method~\cite{gao2022extended} {under the 30-view and 60-view settings.} }

{Moreover, {to further demonstrate the competitiveness of the proposed method, we compare it with four recent state-of-the-art DECT material decomposition methods} on the breast spectral CT dataset: the CLIP-Driven {Universal} Model~\cite{wang2025Clip}, the Dual-Domain Joint Learning Reconstruction Method (JLRM)~\cite{ma2025dual}, the Self-supervised Learning Approach (NoL-MBMI)~\cite{liu2024material}, and the Unrolled ADMM Approach~\cite{10230700}. {For the CLIP-Driven Universal Model, the Unrolled ADMM Approach, and NoL-MBMI, we use the published results obtained under the 520-view setting.} {Specifically, the source code for the CLIP-Driven Universal Model and the Unrolled ADMM Approach} is not publicly available, {while} the NoL-MBMI study reports results under the same 520-view configuration. To {ensure} a fair comparison, the proposed DECT-DRNet is evaluated {using} the same projection-view {setting}, dataset partitioning {scheme}, and data augmentation strategy as {those adopted} in the original studies. DECT-DRNet is independently trained and evaluated three times, and the average performance over these runs is reported. For JLRM, we additionally evaluate its performance at 15, 30, 60, 90, and 520 projection views. The performance of all methods is quantitatively evaluated using peak signal-to-noise ratio (PSNR), structural similarity index (SSIM), and root mean square error (RMSE).} 

{The experiments presented in this paper are implemented using the PyTorch framework.} {The proposed DECT-DRNet is trained using the Adam optimizer with a learning rate of $1 \times 10^{-4}$, a batch size of 4, and 300 training epochs.} The parameters $\eta^k$ and $\rho^k$ are {initialized to} 0.5 and 0.01, respectively. We begin by modifying the architecture of the pretrained U-Net \cite{10.1007/978-3-319-24574-4_28}. Specifically, {the numbers of output channels along} the downsampling path are set to 32, 64, 128, 256, and 512 from the top layer to the bottom {layer.} {The} channel numbers {along} the upsampling path are adjusted accordingly, while all other architectural components remain unchanged. Next, we fine-tune this modified U-Net on the given dataset and then integrate it into the iterative optimization procedure.

{To ensure a fair comparison, the five classical {deep-learning-based} DECT material decomposition methods are retrained on the same training dataset, and all other experimental configurations are kept identical unless explicitly stated otherwise. {The number of training epochs for each of the five methods is set to 500 based on empirical convergence behavior.} Regarding the loss functions, GECCU-Net employs {a} hybrid loss, while the remaining methods adopt the mean squared error {(MSE)} loss, as these configurations provide the best empirical performance in our experiments.} During training, we apply two types of data augmentation {to all these methods: random horizontal and vertical flips.} {For the primal-dual method, the maximum number of iterations is set to $10^6$,  {also based} on empirical convergence observations.}

\begin{table}[!t]
\centering
\caption{{Evaluation of Key Components.}}
\label{tab1}
\setlength{\tabcolsep}{2.mm}{
\begin{tabular*}{0.95\hsize}{@{}@{\extracolsep{\fill}}ccccc@{}}      
\toprule[1.5pt]
\multirow{2}{*}{Method} & Adipose           & Fibroglandular   & Calcification & \multirow{2}{*}{mPSNR} \\
\cmidrule(lr){2-2} \cmidrule(lr){3-3} \cmidrule(lr){4-4}
 & PSNR              & PSNR          & PSNR \\
\midrule
A     & 15.68 & 16.00 & 44.46 & 25.38 \\
B& 20.02&20.27& 48.62 &29.64\\
C& 19.97&20.25& 48.21 &29.48\\
D&20.29&20.50& 48.43 &29.74\\
E&22.10&22.15& 57.71 &33.99\\
F&16.14&16.32&52.06&28.17\\
G&14.18&14.12&49.96&26.09\\
H&{\bf 22.57}&{\bf 22.59}& 54.27 &33.14\\
I& 22.26 &  22.30 & {\bf 57.91} & {\bf 34.16} \\
\bottomrule
\end{tabular*}}
\begin{minipage}{0.95\hsize}
\vspace{4pt}   
\tiny
{A: FBP-JLGD B: Standard CNN regularizer; 
C: Pure image-domain CNN;
D: image-domain CNN with simple FFT; 
E: image-domain CNN with random frequency convolution; 
F: FFC residual block in~\cite{10376896};
G: Band-pass Fourier convolution block in~\cite{10.1007/978-3-031-43999-5_24};
H: proposed model without the FBP-based Jacobian approximation; 
I: DECT-DRNet.}
\end{minipage}
\end{table}

\begin{table}[!t]
\centering
\caption{Evaluation of the number of Fourier residual convolution blocks $l$ in $\mathcal{G}(\cdot)$ and $\tilde{\mathcal{G}}$ in on model performance.}
\label{tab2}
	\setlength{\tabcolsep}{2.mm}{
		\begin{tabular*}{0.95\hsize}{@{}@{\extracolsep{\fill}}
				ccccc@{}}      
		\toprule[1.5pt]
\multirow{2}{*}{$l$} & Adipose & Fibroglandular & Calcification & \multirow{2}{*}{mPSNR} \\
\cmidrule(lr){2-2} \cmidrule(lr){3-3} \cmidrule(lr){4-4}
 & PSNR & PSNR &  PSNR &  \\
		\midrule[0.8pt]
0    & 34.79 &34.57
&59.40&42.92  \\
1    & 38.25&38.16& 61.88
&46.10 \\
2  & 39.83 &39.82&63.56&47.74  \\
3  & 39.84&39.75&67.08&48.89   \\
4 & 40.43&40.36&66.61&49.13   \\
5 & 40.76&40.60&65.76&49.04   \\
\bottomrule[1.5pt]
\end{tabular*}}
\end{table}

\begin{table}[!t]
\centering
\caption{{Evaluation of the number of iterations $K$.}}
\label{tab3}
	\setlength{\tabcolsep}{2.mm}{
		\begin{tabular*}{0.95\hsize}{@{}@{\extracolsep{\fill}}
				ccccc@{}}      
		\toprule[1.5pt]
\multirow{2}{*}{{$K$}} & Adipose & Fibroglandular & Calcification & \multirow{2}{*}{mPSNR} \\
\cmidrule(lr){2-2} \cmidrule(lr){3-3} \cmidrule(lr){4-4}
 & PSNR & PSNR &  PSNR &  \\
\midrule[0.8pt]
1   &22.96
 &22.97
&58.39
&34.77  \\
3    & 36.40&36.28&61.34&44.67\\
5 & 38.90&38.82&{\bf 65.36}&47.69   \\
 7  & {\bf 39.83}&{\bf 39.82}&63.56&{\bf 47.74}   \\
\bottomrule[1.5pt]
\end{tabular*}}
\end{table}

\begin{table}[!ht]
\centering
\caption{{Evaluation of the loss weight parameter $\gamma$.}}
\label{tab4}
\setlength{\tabcolsep}{2.mm}{
\begin{tabular*}{0.95\hsize}{@{}@{\extracolsep{\fill}}ccccc@{}}      
\toprule[1.5pt]
\multirow{2}{*}{$\gamma$} & Adipose & Fibroglandular & Calcification & \multirow{2}{*}{mPSNR} \\
\cmidrule(lr){2-2} \cmidrule(lr){3-3} \cmidrule(lr){4-4}
 & PSNR & PSNR &  PSNR &  \\
\midrule[0.8pt]
0  &39.19& 39.13 & 60.93 & 46.42  \\
0.01 &{\bf 39.83}& {\bf 39.82}&{\bf 63.56}&{\bf 47.74}\\
0.1 & 39.74&39.55 & 60.83 &46.71 \\
1  & 37.94 & 37.83& 58.89 & 44.89   \\
\bottomrule[1.5pt]
\end{tabular*}}
\end{table}

\subsection{Intra-Method Evaluation}
{On the breast spectral CT dataset,} we conduct a series of experiments to evaluate how different configurations of DECT-DRNet affect sparse-view DECT material decomposition. The experiments {investigate} the contributions of key components, the number $l$ of Fourier residual convolutional blocks in $\mathcal{G}(\cdot)$ and $\tilde{\mathcal{G}}(\cdot)$, the number $K$ of iterations, the loss weight $\gamma$, and the {choice} of initial input. All experiments employ $30$ projection views for both training and testing.

\subsubsection{Ablation Study of Key Components}
{To evaluate the contribution of each key component, we compare the proposed DECT-DRNet with several ablated variants, as shown in Table~\ref{tab1}. For this ablation study, {the number of iterations} is fixed at $K=1$}. First, to investigate the effect of the sparse regularization term, we compare the proposed DECT-DRNet with a direct gradient descent approach that uses only the iteration scheme (\ref{eq5}). {This approach, referred to as the FBP-based Jacobian learning gradient descent method (FBP-JLGD), approximates the adjoint Jacobian operator ${\mathrm{D}_k^T(\cdot)}$ using {a} U-Net and FBP. As shown in Table~\ref{tab1}}, DECT-DRNet achieves higher PSNR values than FBP-JLGD for {adipose}, {fibroglandular}, and {calcification} tissue maps, which demonstrates that the framework with the sparse regularization term effectively enhances model performance. {Moreover, when compared with several alternative variants that modify key components such as the standard CNN regularizer, a purely image-domain CNN, an image-domain CNN combined with a simple FFT, an image-domain CNN with random frequency convolution, the FFC residual block, and the band-pass Fourier convolution block, our proposed model achieves superior performance. This demonstrates that the designed frequency-domain convolution effectively improves reconstruction quality. In addition, relative to the variant without the FBP-based Jacobian approximation, our model further improves the PSNR of the {calcification} tissue map as well as the mPSNR, confirming the effectiveness of the proposed FBP-based Jacobian learning {module.} Overall, all key components make positive contributions to the final performance.}

\begin{figure*}[!ht]
\centering
\includegraphics[width=0.9\linewidth]{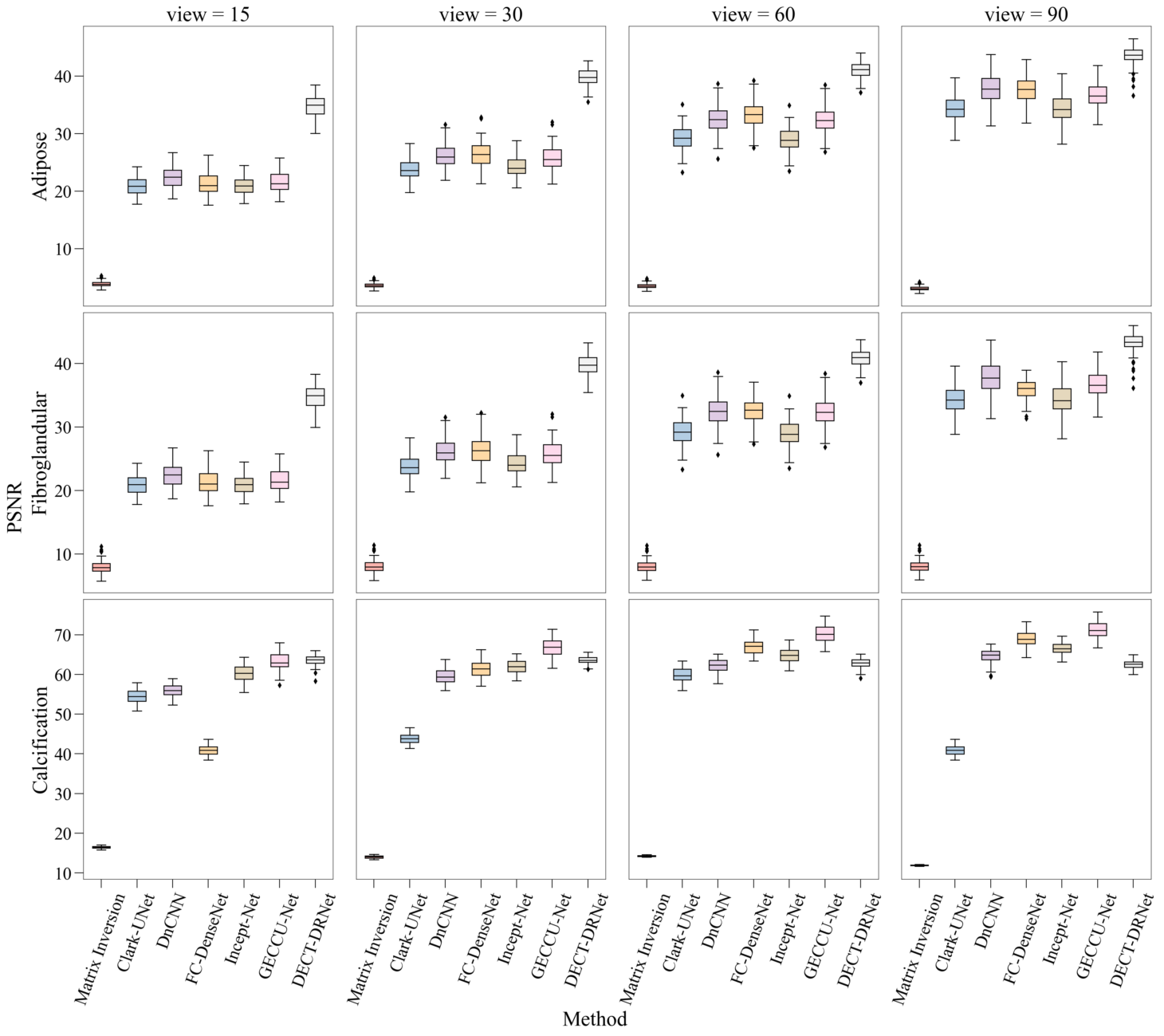}
\caption{Boxplots of PSNR values of different methods.} \label{fig3}
\end{figure*}

\begin{table}[!ht]
\centering
\caption{Evaluation of different initial values on model performance.}
\label{tab5}
\setlength{\tabcolsep}{2.mm}{
\begin{tabular*}{0.95\hsize}{@{}@{\extracolsep{\fill}}ccccc@{}}      
\toprule[1.5pt]
Initialization & Adipose & Fibroglandular & Calcification & \multirow{2}{*}{mPSNR} \\
\cmidrule(lr){1-1} \cmidrule(lr){2-2} \cmidrule(lr){3-3} \cmidrule(lr){4-4}
$\boldsymbol{x}_0$ & PSNR & PSNR &  PSNR &  \\
\midrule[0.8pt]
$\boldsymbol{0}$   &38.79 & 38.60 & 58.11 & 45.17  \\
$\text{U-Net} \circ \text{FBP}(y)$  & {\bf 39.83} &{\bf 39.82}  & {\bf 63.56} &{\bf 47.74}  \\
\bottomrule[1.5pt]
\end{tabular*}}
\end{table}

\begin{table*}[!t]
\centering
\caption{Quantitative decomposition results of different methods on the breast
dataset.}
\label{tab6}
\setlength{\tabcolsep}{2.mm}{
\begin{tabular*}{0.98\hsize}{@{}@{\extracolsep{\fill}}clcccccccc@{}}      
\toprule[1.5pt]
\multirow{2}{*}{View} & \multirow{2}{*}{Method} & \multicolumn{2}{c}{Adipose} & \multicolumn{2}{c}{Fibroglandular} & \multicolumn{2}{c}{Calcification} &\multirow{2}{*}{mPSNR} &\multirow{2}{*}{mSSIM}\\
\cmidrule(lr){3-4} \cmidrule(lr){5-6} \cmidrule(lr){7-8}
& & PSNR & SSIM & PSNR & SSIM & PSNR & SSIM \\
\midrule[0.8pt]

\multirow{7}{*}{15} 
& Matrix Inversion    & 3.91  &0.0199 &7.93
  &0.2697
  &16.47  & 0.0690&9.44&0.1195 \\

& Clark-UNet~\cite{Clark2018MultienergyCD}  & 20.92 &0.8808& 20.91 & 0.8794&54.47  & 0.9991&32.10&0.9198\\
&DnCNN~\cite{lu2019learning}        &\underline{22.46} & 0.9159& \underline{22.46} & 0.8984& 55.98   &0.9967 &33.63&0.9370\\
& FC-DenseNet~\cite{wu2019multi}           & 21.33&0.9321  & 21.33 & 0.9316 &40.92& 0.9942&27.86&0.9526\\
& Incept-Net~\cite{https://doi.org/10.1002/mp.14523}  & 21.01 & 0.8662 & 21.00 &0.7442& 60.24  &0.9990&34.08&0.8698\\

&GECCU-Net~\cite{shi2024multi}  & 21.62 &\underline{0.9360}&21.62 & \underline{0.9356} & \underline{63.22}& \textbf{0.9999} &\underline{35.48}&\underline{0.9572}\\
&DECT-DRNet & \textbf{34.74} &\textbf{0.9880} &  \textbf{34.67} &\textbf{0.9885}&\textbf{63.61}&  \underline{0.9998} & \textbf{44.34}& \textbf{0.9921}\\
\midrule[0.8pt]

\multirow{7}{*}{30} 
& Matrix Inversion    & 3.67 & 0.0102 & 8.04 & 0.3481 & 14.04 & 0.0346 &8.58&0.1310\\

& Clark-UNet~\cite{Clark2018MultienergyCD}  & 23.82 & 0.9310 &23.82  &0.9215  & 43.85& 0.9970& 30.50&0.9498\\

&DnCNN~\cite{lu2019learning}        &26.09 & 0.9490 & 26.09 & 0.8844 & 59.66 & 0.9992 &37.28&0.9442\\
& FC-DenseNet~\cite{wu2019multi}           & \underline{26.28} & \underline{0.9667}&\underline{26.11}   & \underline{0.9659} & 61.31  & \textbf{0.9999} &37.90&\underline{0.9775}\\
& Incept-Net~\cite{https://doi.org/10.1002/mp.14523}  & 24.23 & 0.8999 & 24.22 & 0.9109 & 61.92 & 0.9984&36.79&0.9364\\

&GECCU-Net~\cite{shi2024multi}  & 25.78 & 0.9642 &25.78 & 0.9639& \textbf{66.86} & \textbf{0.9999} & \underline{39.47}&0.9760\\
&DECT-DRNet &\textbf{39.83} & \textbf{0.9943} & \textbf{39.82} & \textbf{0.9942}& \underline{63.56} & \underline{0.9997} &\textbf{47.74}  &\textbf{0.9961}\\

\midrule[0.8pt]
\multirow{7}{*}{60} 
& Matrix Inversion    &3.57& 0.0139 & 8.05 & 0.2791& 14.29 & 0.0555 &8.63&0.1162\\

& Clark-UNet~\cite{Clark2018MultienergyCD}  & 29.30 & 0.9719 & 29.29 &0.9721 &59.88 & 0.9998&39.49&0.9813 \\
& DnCNN~\cite{lu2019learning}        & 32.53 & 0.9808 & 32.52 & 0.9825 &62.32&0.9995&42.46&0.9876\\
& FC-DenseNet~\cite{wu2019multi}         & \underline{33.28}&  \underline{0.9899}& \underline{32.55}&\underline{0.9896}& \underline{66.98}& \textbf{0.9999} &44.27&\underline{0.9931}\\
& Incept-Net~\cite{https://doi.org/10.1002/mp.14523} & 29.03 & 0.9587&29.03 & 0.9587 & 64.84& \underline{0.9998} &40.97&0.9724\\
& GECCU-Net~\cite{shi2024multi}& 32.38& 0.9886 & 32.37& 0.9883 & \textbf{70.32} & \textbf{0.9999} &\underline{45.02}&0.9923\\
& DECT-DRNet&\textbf{41.07} &\textbf{0.9946} & \textbf{40.91} & \textbf{0.9960} & 62.89 & 0.9997 & \textbf{48.29}& \textbf{0.9968}\\

\midrule[0.8pt]
\multirow{7}{*}{90} 
& Matrix Inversion    &3.13& 0.0102 &8.08 & 0.3471 & 11.93 & 0.0392 &7.71&0.1322\\
& Clark-UNet~\cite{Clark2018MultienergyCD}  & 34.37 & 0.9887 & 34.33 &0.9891 &40.92 & 0.9936 &36.54&0.9905\\
& DnCNN~\cite{lu2019learning}        & \underline{37.75}& 0.9868 & \underline{37.74} & 0.9928 &64.74 &0.9995 &46.74&0.9930\\
& FC-DenseNet~\cite{wu2019multi}         & 37.60&0.9950& 35.92&\underline{0.9952}& \underline{69.00}& \textbf{0.9999} &47.51&0.9967\\
& Incept-Net~\cite{https://doi.org/10.1002/mp.14523} & 34.43 & 0.9857 & 34.40 & 0.9785 & 66.55 & \underline{0.9996}&45.13&0.9879\\
& GECCU-Net~\cite{shi2024multi}& 36.66 & \underline{0.9952} & 36.67& \underline{0.9952} & \textbf{71.21} & \textbf{0.9999} &\underline{48.18}& \underline{0.9968}\\
& DECT-DRNet&\textbf{43.50} &\textbf{0.9961} &\textbf{43.21} & \textbf{0.9962} & 62.58 & 0.9991 & \textbf{49.76}&\textbf{0.9971}\\

\bottomrule[1.5pt]
\end{tabular*}
}
\end{table*}

\subsubsection{Number of Fourier Residual Convolutional Blocks $l$ in $\mathcal{G}(\cdot)$ and $\tilde{\mathcal{G}}(\cdot)$}
To evaluate {how} incorporating information from the geometric transformation field $\mathcal{T}_\mathcal{G}=\{\boldsymbol{\phi}\in\mathbb{R}^q|\boldsymbol{\phi}=\mathcal{G}(\boldsymbol{x})\}$ {affects} decomposition performance, we analyze the relationship
between the number of Fourier residual {convolutional} blocks
and the model’s decomposition accuracy. In {the} experiments, {the number of iterations} is {fixed}  at ${K}=7$. Meanwhile, {the number  $l$ of Fourier residual convolutional blocks  is varied from 0 to 5 in increments of 1.} As shown in Table~\ref{tab2}, models {incorporating} Fourier residual convolution blocks {achieve} better decomposition results {than} those {without} these {blocks}, {demonstrating} that {frequency-domain} information can effectively enhance model performance. Additionally, the mean PSNR (mPSNR) {across} the three tissue maps {increases with the number of Fourier residual convolutional blocks} and tends to stabilize when $l \geq 2$. Therefore, to balance decomposition accuracy and computational efficiency, we {set the number of} Fourier residual {convolutional} blocks {to} $l=2$ in our subsequent experiments.

\subsubsection{Number of Iterations $K$}
The number of iterations is a key parameter in iterative unrolling models. In this part, we {investigate the effect of the number of iterations} {$K$} {on} the decomposition performance of the proposed model. Table~\ref{tab3} presents the experimental results of PSNR for {the} three tissue maps {as the number of iterations $K$ increases} from 1 to 7 {in increments of} 2. {As shown in} Table~\ref{tab3}, {the mPSNR generally increases with the number of iterations, indicating improved decomposition performance and supporting the effectiveness of the iterative unrolling design.} Therefore, we set ${K}=7$ in subsequent experiments.

\subsubsection{Analysis of Different Loss {Weights} $\gamma$}
We perform experiments with DECT-DRNet to assess how the weight parameter $\gamma$ in the loss function affects the performance of the model. {Specifically, we test $\gamma$ values of 0, 0.01, 0.1, and 1.} As indicated in Table~\ref{tab4},  $\gamma=0.01$ yields the best decomposition performance for {the} {adipose}, {fibroglandular}, and {calcification} tissue maps. Therefore, we choose $\gamma = 0.01$  to ensure optimal model performance.

\begin{figure*}[!ht]
\centering
\includegraphics[width=0.93\linewidth]{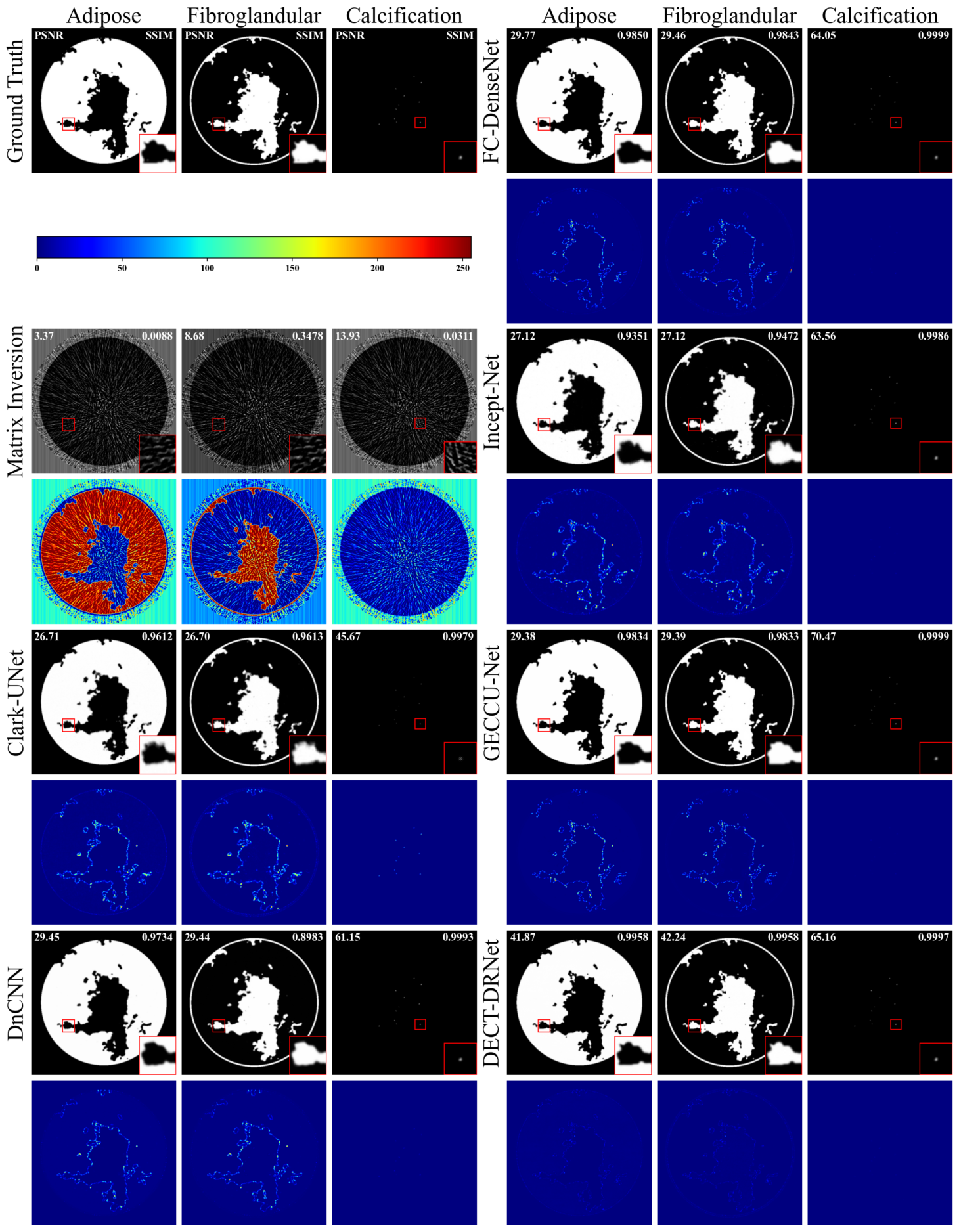}
\caption{Decomposition results of different methods and corresponding residual maps on breast spectral CT with 30 projection views.} \label{fig4}
\end{figure*}

\subsubsection{Different Initial Values}
To evaluate the stability of our proposed model, we compare two initialization strategies: $\boldsymbol{x}_0 = \boldsymbol{0}$ and 
$\boldsymbol{x}_0 = \text{U-Net} \circ \text{FBP}(\boldsymbol{y})$. We evaluate the  material decomposition performance {under both strategies.} As shown in Table~\ref{tab5}, the results for  the {adipose}, {fibroglandular}, and {calcification} tissue maps remain generally consistent across the two initialization strategies. These results {indicate} that the model's performance is largely {insensitive to} the choice of initialization, {demonstrating} its {robustness to different initialization strategies.} Furthermore, {initialization with} $\boldsymbol{x}_0 = \boldsymbol{0}$ {yields} slightly {lower decomposition performance than initialization with} $\boldsymbol{x}_0 = \text{U-Net} \circ \text{FBP}(\boldsymbol{y})$, {supporting} the effectiveness of the proposed initialization strategy.

\begin{table*}[!ht]
\centering
\caption{{Quantitative decomposition results of different methods on the noisy breast 
dataset.} }
\label{tab7}
\setlength{\tabcolsep}{1.mm}{
\begin{tabular*}{0.98\hsize}{@{}@{\extracolsep{\fill}}clcccccccc@{}}      
\toprule[1.5pt]
\multirow{2}{*}{View} & \multirow{2}{*}{Method} & \multicolumn{2}{c}{Adipose} & \multicolumn{2}{c}{Fibroglandular} & \multicolumn{2}{c}{Calcification} &\multirow{2}{*}{mPSNR} &\multirow{2}{*}{mSSIM}\\
\cmidrule(lr){3-4} \cmidrule(lr){5-6} \cmidrule(lr){7-8} 
& & PSNR & SSIM & PSNR & SSIM & PSNR & SSIM \\
\midrule[0.8pt]

\multirow{7}{*}{15} 

& Clark-UNet  & $20.84 \pm 1.53$ &$0.8806 \pm 0.0316$& $20.83  \pm 1.52$ & $0.8799  \pm 0.0360$&$54.50 \pm 1.73$ & $0.9991 \pm 0.0003$&32.06 &0.9199\\
&DnCNN        &$20.19\pm1.65$ & $0.8781\pm 0.0358$& $20.19\pm1.65$ & $0.8884\pm0.0303$& $55.96\pm1.46$   &$0.9967\pm0.0007$ &32.11&0.9211\\
& FC-DenseNet           & $21.32\pm1.87$ &$0.9320\pm 0.0237$ & $21.32\pm1.87$ & $0.9315\pm0.0232$ &$40.92\pm 1.20$& $0.9942\pm0.0015$&27.85&0.9526 \\
& Incept-Net  & $18.09\pm1.45$& $0.8297\pm0.0428$ & $18.09\pm1.45$ &$0.6689\pm0.0278$& $60.06\pm2.10$  &$0.9989\pm0.0002$ &32.08&0.8325\\

&GECCU-Net  & $21.58\pm1.73$ &$0.9356\pm0.0211$&$21.58\pm1.76$ & $0.9353\pm0.0210$&$63.18\pm2.30$& $0.9999\pm 0.0000$ &\underline{35.45}&\underline{0.9569}\\
&DECT-DRNet &$34.64\pm1.84$ &$0.9877\pm0.0051$ & $34.58\pm 1.85$ &$0.9882\pm 0.0048$&$63.60\pm0.94$&  $0.9998\pm0.0000$ & \textbf{44.27}& \textbf{0.9919}\\
\midrule[0.8pt]

\multirow{7}{*}{30} 

& Clark-UNet  & $23.81 \pm 1.74$ & $0.9310\pm 0.0232$ &$23.81\pm 1.73$  &$0.9216\pm 0.0293$  & $43.85\pm 1.20$& $0.9970\pm 0.0008$& 30.49&0.9499\\
&DnCNN        &$24.76\pm1.85$& $0.9243\pm0.0266$ & $24.75\pm1.85$ & $0.8800\pm0.0116$ & $59.80\pm1.70$ & $0.9992\pm 0.0001$ &36.44&0.9345\\
& FC-DenseNet           & $26.27\pm2.18$ & $0.9667\pm0.0141$&$26.10\pm2.10$   & $0.9658\pm0.0139$ & $61.31\pm 1.87$  & $0.9999\pm 0.0000$ &37.89&\underline{0.9775}\\
& Incept-Net  & $22.50\pm 1.55$& $0.8694\pm0.0325$ & $22.50\pm1.55$ & $0.8841\pm0.0283$ & $61.39\pm1.41$& $0.9977\pm0.0003$ &35.46&0.9171\\

&GECCU-Net  & $25.73\pm2.04$ & $0.9639\pm0.0138$ &$25.73\pm2.04$ & $0.9636\pm 0.0135$& \textbf{$66.82\pm2.30$} & \textbf{$0.9999\pm 0.0000$} & \underline{39.43}&0.9758\\
&DECT-DRNet &\textbf{$39.65\pm1.51$} & \textbf{$0.9942\pm0.0014$} & \textbf{$39.63\pm1.64$} & \textbf{$0.9941\pm0.0013$}& $63.54\pm0.94$ &$0.9997\pm0.0000$ &\textbf{47.61}  &\textbf{0.9960}\\

\midrule[0.8pt]
\multirow{7}{*}{60} 

& Clark-UNet  & $29.29\pm 2.07$ & $0.9718\pm 0.0167$ & $29.28\pm2.05$ & $0.9720\pm0.0168$ &$59.88\pm1.70$ & $0.9998\pm 0.0001$&39.48&0.9812 \\
& DnCNN        & $32.18\pm2.21$ & $0.9797\pm0.0085$ & $32.17\pm2.21$ & $0.9817\pm0.0074$ &$62.32\pm1.49$&$0.9995\pm0.0000$&42.22&0.9870 \\
& FC-DenseNet         &$33.26\pm2.19$& $0.9898\pm0.0053$& $32.54\pm1.88$&$0.9896\pm0.0051$& $66.98\pm1.81$& \textbf{$0.9999\pm0.0000$} &44.26&\underline{0.9931}\\
& Incept-Net & $28.48\pm 1.95$ & $0.9553\pm0.0216$ &$28.50\pm1.95$ & $0.9586\pm0.0206$ & $64.84\pm1.82$ & $0.9998\pm0.0000$ &40.61&0.9712\\
& GECCU-Net& $32.36\pm2.19$ & $0.9885\pm0.0057$ & $32.35\pm2.19$& $0.9883\pm0.0057$ & \textbf{$70.27\pm2.09$} & \textbf{$0.9999\pm0.0000$} &\underline{45.00}&0.9922\\
& DECT-DRNet&\textbf{$40.98\pm1.40$} &\textbf{$0.9945\pm0.0013$} & \textbf{$40.82\pm1.37$} & \textbf{$0.9959\pm0.0010$} & $62.89\pm1.14$ & $0.9997\pm0.0001$ & \textbf{48.23}& \textbf{0.9967}\\

\midrule[0.8pt]
\multirow{7}{*}{90} 
& Clark-UNet  & $34.35\pm2.16$ & $0.9887\pm0.0068$ & $34.31\pm2.14$ &$0.9890\pm0.0075$ &$40.92\pm1.20$ & $0.9936\pm0.0016$ &36.53&0.9904\\
& DnCNN        & $37.56\pm2.32$ & $0.9865\pm0.0050$ & $37.54\pm2.32$ & $0.9926\pm 0.0036$ &$64.71 \pm1.61$&$0.9995\pm 0.0001$ &46.60&0.9929\\
& FC-DenseNet         & $37.57\pm2.01$&$0.9950\pm0.0023$& $35.89\pm1.44$&$0.9951\pm0.0023$& $68.99\pm1.94$& \textbf{$0.9999\pm0.0000$} &47.48& \underline{0.9967}\\
& Incept-Net & $34.29\pm2.32$ & $0.9856\pm0.0100$ & $34.26\pm2.31$ & $0.9784\pm0.0111$ & $66.51\pm1.51$ & $0.9996\pm 0.0000$ &45.02&0.9879\\
& GECCU-Net& $36.63\pm2.05$ & $0.9952\pm0.0023$ & $36.63\pm2.05$& $0.9951\pm0.0022$ & \textbf{$71.20\pm2.10$} & \textbf{$0.9999\pm0.0000$} &\underline{48.15}& \underline{0.9967}\\
& DECT-DRNet&\textbf{$43.38\pm1.61$} &\textbf{$0.9961\pm0.0012$} &\textbf{$43.10\pm 1.63$} & \textbf{$0.9961\pm0.0008$}& $62.57\pm 0.92$ & $0.9991\pm0.0003$ & \textbf{49.68}& \textbf{0.9971}\\

\bottomrule[1.5pt]
\end{tabular*}
}
\end{table*}

\begin{table}[!t]
\centering
\caption{{Quantitative decomposition results of different methods on the breast dataset under multiple noise levels.}}
\label{tab8}
\setlength{\tabcolsep}{0.68mm}{
\begin{tabular*}{0.98\hsize}{@{}@{\extracolsep{0pt}}clcccc@{}}
\toprule[1.5pt]
\multirow{2}{*}{Noise level} & \multirow{2}{*}{Method} & {Adipose} & {Fibroglandular} & {Calcification} & \multirow{2}{*}{mRMSE}\\
\cmidrule(lr){3-3} \cmidrule(lr){4-4} \cmidrule(lr){5-5}
& & RMSE & RMSE & RMSE \\
\midrule[0.8pt]
\multirow{6}{*}{Level 1}
& Clark-UNet  & 0.0658 & 0.0658 & 0.0065 & 0.0460 \\
& DnCNN       & 0.0591 & 0.0592 & 0.0010 & 0.0398 \\
& FC-DenseNet & \underline{0.0501} & \underline{0.0509} & 0.0009 & \underline{0.0340} \\
& Incept-Net  & 0.0762 & 0.0762 & 0.0009 & 0.0511 \\
& GECCU-Net   & 0.0531 & 0.0531 & \textbf{0.0005} & 0.0356 \\
& DECT-DRNet  & \textbf{0.0106} & \textbf{0.0106} & \underline{0.0007} & \textbf{0.0073} \\
\midrule[0.8pt]
\multirow{6}{*}{Level 2}
& Clark-UNet  & 0.0661 & 0.0661 & 0.0065 & 0.0462 \\
& DnCNN       & 0.0912 & 0.0912 & 0.0011 & 0.0612 \\
& FC-DenseNet & \underline{0.0502} & \underline{0.0511} & 0.0009 & \underline{0.0341} \\
& Incept-Net  & 0.1295 & 0.1293 & 0.0011 & 0.0866 \\
& GECCU-Net   & 0.0540 & 0.0540 & \textbf{0.0005} & 0.0362 \\
& DECT-DRNet  & \textbf{0.0112} & \textbf{0.0113} & \underline{0.0007} & \textbf{0.0077} \\
\midrule[0.8pt]
\multirow{6}{*}{Level 3}
& Clark-UNet  & 0.0691 & 0.0691 & 0.0065 & 0.0482 \\
& DnCNN       & 0.1476 & 0.1476 & 0.0013 & 0.0988 \\
& FC-DenseNet & \underline{0.0511} & \underline{0.0520} & 0.0009 & \underline{0.0347} \\
& Incept-Net  & 0.2788 & 0.2783 & 0.0020 & 0.1864 \\
& GECCU-Net   & 0.0610 & 0.0610 & \textbf{0.0006} & 0.0409 \\
& DECT-DRNet  & \textbf{0.0156} & \textbf{0.0157} & \underline{0.0007} & \textbf{0.0107} \\
\bottomrule[1.5pt]
\end{tabular*}
}
\end{table}

\begin{figure*}[!t]
\centering
\includegraphics[width=0.92\linewidth]{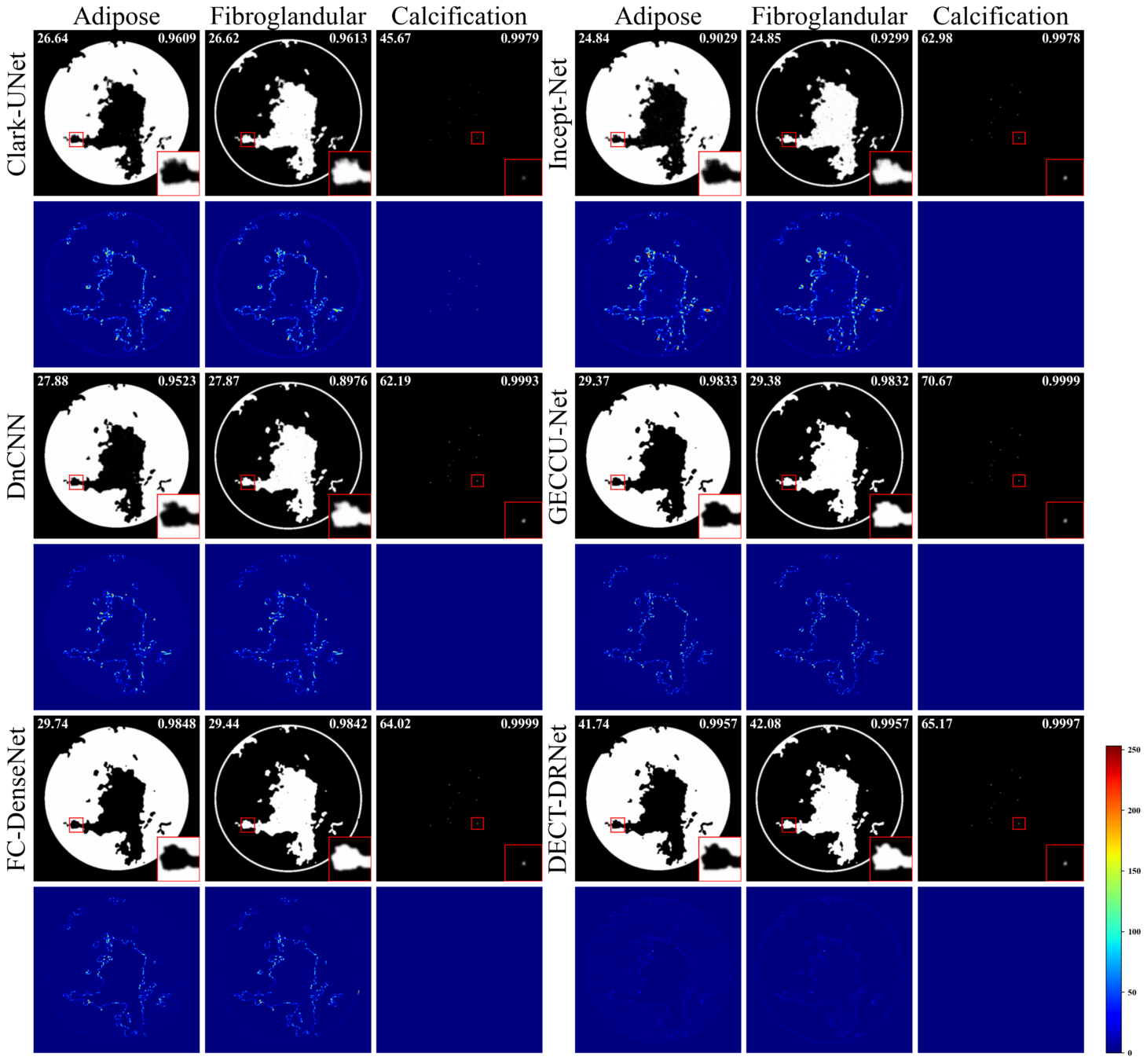}
\caption{Decomposition results of different methods and corresponding residual maps on noisy breast spectral CT with 30 projection views.} \label{fig5}
\end{figure*}

\subsection{Comparisons with Classical Deep Learning-Based Methods}
\subsubsection{Breast Spectral CT Dataset}
 Table~\ref{tab6} {presents} the PSNR and SSIM values obtained by these methods for the {adipose}, {fibroglandular}, and {calcification} tissue maps {under settings of} 15, 30, 60, and 90 projection views. The results indicate that DECT-DRNet consistently outperforms all competing methods in decomposing {adipose} and {fibroglandular} tissue maps under various sparse-view scenarios. Specifically, {at} 15, 30, 60, and 90 projection views, {the corresponding PSNR improvements are} 12.28 dB, 13.55 dB, 7.79 dB, and 5.75 dB for the {adipose} tissue map, and 12.21 dB, 13.71 dB, 8.36 dB, and 5.47 dB for the {fibroglandular} tissue map. Furthermore, DECT-DRNet achieves the highest mPSNR and mSSIM across the three tissue maps {under all projection-view settings.}

To evaluate the stability of DECT-DRNet, Fig.~\ref{fig3} illustrates boxplots of the experimental results on breast spectral CT data under 15, 30, 60, and 90 projection views. As shown in Fig.~\ref{fig3}, DECT-DRNet achieves the highest lower quartile, median, and upper quartile {PSNR} values for {the} {adipose} and {fibroglandular} tissue maps under different sparse-view conditions. {For the} {calcification} tissue {map}, DECT-DRNet also demonstrates superior stability and performance.

Fig.~\ref{fig4} {presents} the visual decomposition results {obtained by} several methods using 30 projection views. As shown in Fig.~\ref{fig4}, the traditional Matrix Inversion decomposition method {produces severe artifacts} in all three tissue maps and fails to reconstruct their fundamental structures. Although Clark-UNet and Incept-Net manage to recover the basic structures in {adipose} and {fibroglandular} tissue maps, they still exhibit noticeable artifacts. DnCNN, FC-DenseNet, and GECCU-Net can effectively suppress artifacts and achieve overall image reconstruction, {but} they do not entirely preserve fine details. In contrast, DECT-DRNet not only substantially reduces noise and artifacts but also {preserves} fine structural details, achieving the best decomposition performance for the {adipose} and {fibroglandular} tissue maps. For the {calcification} tissue map, DECT-DRNet effectively resolves these {fine structures,} with its performance surpassed only by GECCU-Net.

To further verify the {detail-preservation capability} of DECT-DRNet, Fig.~\ref{fig4} {shows} the absolute difference maps between each method's predictions and the ground truth for the three tissue maps {under the 30-view setting}. {Compared with the other methods,} DECT-DRNet exhibits the {lowest} residual errors for the {adipose} and {fibroglandular} tissue maps, indicating better detail preservation. DECT-DRNet also effectively {preserves} details in {the} {calcification} tissue map.

\subsubsection{Examination of Robustness Against Noise}
To assess the robustness of the proposed DECT-DRNet, Gaussian noise with a standard deviation of $5\times{10^{-6}}$ {is added to} the projection data {in} the test set. The trained network {is evaluated on the noisy test data} without further retraining. Table~\ref{tab7} presents {the decomposition results of various methods for {the} three tissue maps under different sparse-view settings.} Compared with the corresponding noise-free results in Table~\ref{tab6}, DECT-DRNet maintains comparable PSNR and SSIM values and achieves the highest mPSNR and mSSIM in the three tissue maps. {These results demonstrate} that DECT-DRNet {is robust to} Gaussian noise. Fig.~\ref{fig5} provides a visual {comparison} of {the results obtained by each method under the 30-view setting, with Gaussian noise added to the projection data.} {The results in} Fig.~\ref{fig5} show that Gaussian noise {introduces additional artifacts in the results obtained by Incept-Net, while some artifacts are also observed in those obtained by DnCNN.} However, the decomposition {results} for Clark-UNet, FC-DenseNet, GECCU-Net, and DECT-DRNet remain mostly stable {under noise conditions,} indicating strong {robustness to noise.} Despite this, the images show that Clark-UNet, FC-DenseNet, and GECCU-Net {preserve less fine structural} detail in the {adipose} and {fibroglandular} tissue maps {than} DECT-DRNet. Additionally, DECT-DRNet performs well in the decomposition of the {calcification} tissue map.

{To examine how {the} noise level affects model performance, we perform additional 30-view experiments by adding Gaussian noise with different standard deviations to the projection data. As summarized in Table~\ref{tab8}, level 1, level 2, and level 3 correspond to Gaussian noise with standard deviations of $5\times10^{-6}$, $1\times10^{-5}$, and $2.5\times10^{-5}$, respectively. The results show that, as the noise level increases, the RMSE values of all evaluated methods generally increase for the three tissue maps. Compared with the other {methods}, DECT-DRNet consistently attains the lowest mRMSE at all three noise levels, with values of 0.0073, 0.0077, and 0.0107, respectively. These results indicate that DECT-DRNet maintains stable decomposition performance {across} different noise {levels}, {thereby} demonstrating strong robustness to noise.}

\subsubsection{Abdominal Dataset}
{
To assess the {generalizability} of the proposed method, we {conduct} additional experiments on an abdominal CT dataset. Table~\ref{tab9} presents a quantitative comparison {between the proposed approach and} five widely adopted deep learning-based DECT material decomposition methods, including Clark-UNet, DnCNN, FC-DenseNet, Incept-Net, and GECCU-Net. As reported in Table~\ref{tab9}, for both 30 and 60 projection views, the proposed method achieves the highest PSNR and SSIM values, demonstrating its superior material decomposition performance on the abdominal CT dataset. Moreover, the relatively small standard deviations {obtained} by the proposed method {indicate} lower performance variability across different test samples and greater overall stability.}

\begin{figure*}[!ht]
\centering
\includegraphics[width=0.95\linewidth]{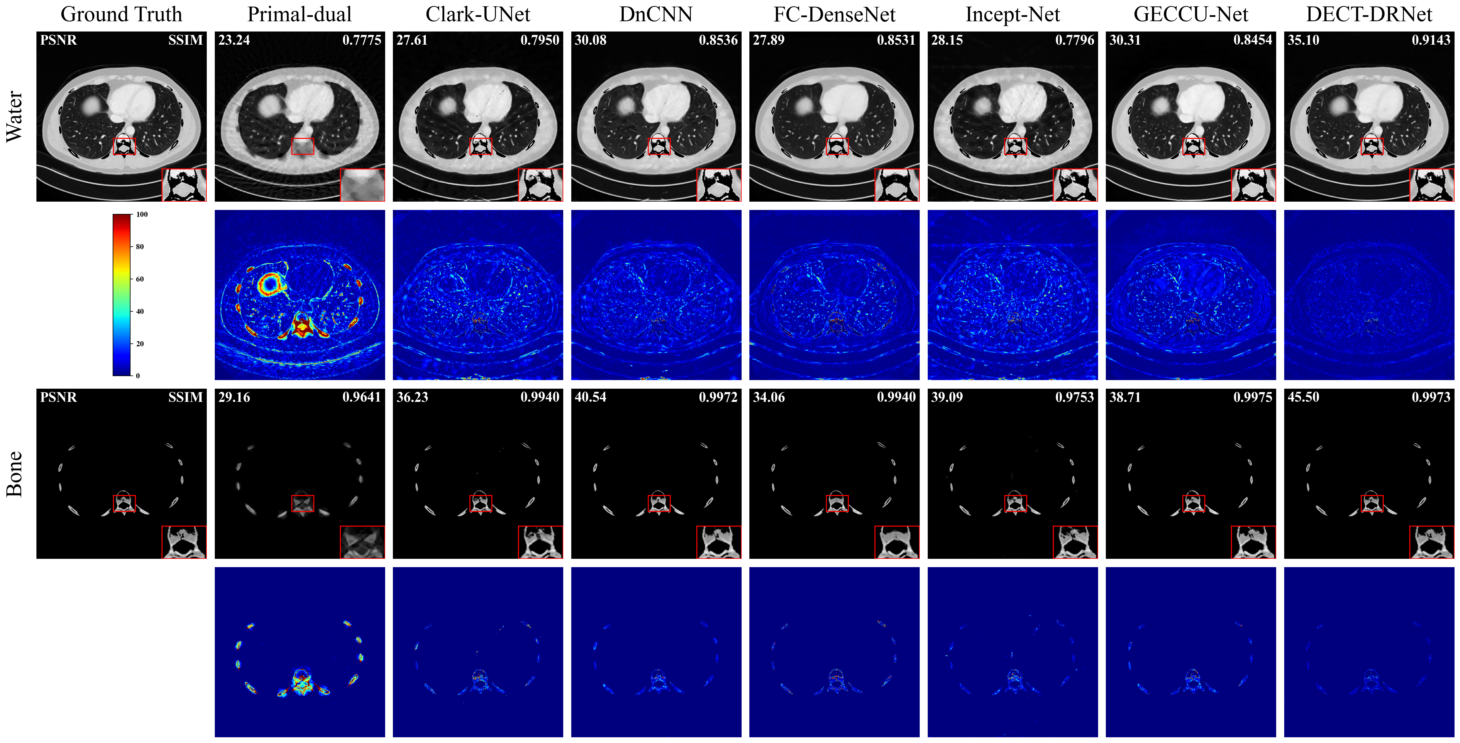}
\caption{{Decomposition results of different methods and corresponding residual maps on the abdominal 
dataset with 30 projection views.}} \label{fig6}
\end{figure*}

{Fig.~\ref{fig6} {presents a} visual comparison of different methods under the 30-view setting, with red rectangles marking {zoomed-in} regions of interest. The model-driven primal-dual approach {preserves} the {overall} structural contours of the water and bone material images but still produces {noticeable} artifacts. Clark-UNet, DnCNN, FC-DenseNet, Incept-Net, and GECCU-Net alleviate artifacts to {some extent but  remain less effective at} restoring fine details and preserving edges. In comparison, the proposed method achieves better detail reconstruction and {artifact suppression}, yielding material decomposition results with more complete structural information and sharper edges.}

\begin{table*}[!t]
\centering
\caption{{Quantitative decomposition results of different methods on the abdominal 
dataset.} }
\label{tab9}
\setlength{\tabcolsep}{2.mm}{
\begin{tabular*}{0.98\hsize}{@{}@{\extracolsep{\fill}}clcccccc@{}}      
\toprule[1.5pt]
\multirow{2}{*}{View} & \multirow{2}{*}{Method} & \multicolumn{2}{c}{Water} & \multicolumn{2}{c}{Bone}  &\multirow{2}{*}{mPSNR} &\multirow{2}{*}{mSSIM}\\
\cmidrule(lr){3-4} \cmidrule(lr){5-6}
& & PSNR & SSIM  & PSNR & SSIM \\
\midrule[0.8pt]

\multirow{7}{*}{30} 

& Clark-UNet  & $30.13 \pm 1.02$ &$0.8241 \pm 0.0135$ &$41.27\pm3.76$ & $0.9973\pm 0.0023$& 35.70&0.9107\\
&DnCNN        &$32.18\pm 0.90$ & $0.8683\pm 0.0100$& $43.39\pm 2.89$   & $0.9975\pm0.0010$  &\underline{37.79}&0.9329\\
& FC-DenseNet           & $30.78\pm1.62$ & $0.8661\pm 0.0120$ &$38.70\pm3.76$ &  $0.9967\pm0.0023$ &34.74&0.9314\\
& Incept-Net  & $30.42\pm0.98$& $0.8191\pm0.0198$ &$43.11\pm3.65$ &$0.9740\pm0.0017$ &36.77&0.8966\\

&GECCU-Net  & $32.41\pm1.21$ & $0.8692\pm0.0135$   & $43.02\pm4.32$& $0.9984\pm 0.0014$ &37.72&\underline{0.9338}\\
& DECT-DRNet & $36.50\pm0.88$ &  $0.9135\pm 0.0090$&$48.27\pm2.84$ & $0.9972\pm0.0005$ &\textbf{42.39}& \textbf{0.9554}\\
\midrule[0.8pt]

\multirow{7}{*}{60} 

& Clark-UNet  & $34.85\pm0.86$ & $0.8854\pm0.0090$ &$46.93\pm3.15$  & $0.9993\pm0.0006$ & 40.89 &0.9424\\
&DnCNN        &$36.19\pm0.76$& $0.9061\pm0.0066$ & $48.49\pm2.99$ & $0.9985\pm0.0005$ &\underline{42.34}&0.9523\\
& FC-DenseNet           & $34.66\pm1.41$ & $0.9061\pm0.0095$& $43.13\pm3.41$   & $0.9987\pm0.0010$  &38.90&0.9524\\
& Incept-Net  & $35.18\pm0.78$& $0.8933\pm0.0075$  &$47.23\pm2.98$ &$0.9977\pm0.0006$&41.21&0.9455\\

&GECCU-Net  & $36.42\pm0.93$& $0.9210\pm0.0069$ & $48.07\pm3.55$& $0.9993\pm0.0006$ & 42.25 & \underline{0.9602} \\
&DECT-DRNet & $38.48\pm0.81$ & $0.9329\pm0.0064$&$50.21\pm2.54$ & $0.9976\pm0.0006$ & \textbf{44.35} & \textbf{0.9653} \\

\bottomrule[1.5pt]
\end{tabular*}
}
\end{table*}

\begin{table*}[!ht]
\centering
\caption{{Quantitative decomposition results of the recent representative methods on the breast 
dataset.}}
\label{tab10}
\vspace{4pt}
\renewcommand{\arraystretch}{0.8}
\setlength{\tabcolsep}{0.3mm}{
\begin{tabular*}{0.98\hsize}{@{}@{\extracolsep{\fill}}lccccc@{}}
\toprule[1.5pt]
Method & \makecell[c]{Train:test \\sample ratio} & Augmentation & View & mRMSE & mPSNR \\
\midrule[0.8pt]
\multirow{5}{*}{JLRM \cite{ma2025dual}} 
& \multirow{5}{*}{8:1} 
& \multirow{5}{*}{\makecell[c]{Horizontal flip\\ Vertical flip}}
& 15  & 0.2198& 18.98\\
& & & 30  & 0.2070 & 19.33 \\
& & & 60  & 0.1946 &  19.71\\
& & & 90  & 0.1908 & 19.78 \\
& & & 520 & 0.0017 & 81.42 \\
\midrule[0.8pt]
\multirow{7}{*}{Our Method} 
& \multirow{5}{*}{8:1} 
& \multirow{5}{*}{\makecell[c]{Horizontal flip\\Vertical flip}} 
& 15  & 0.0129 & 44.34 \\
& & & 30  & 0.0073 & 47.74 \\
& & & 60  & 0.0062& 48.29 \\
& & & 90  & 0.0033 & 49.76 \\
& & & 520 &0.0030  & 52.53 \\
\cmidrule(lr){2-6}
& 8:1 
& \makecell[c]{-} 
& 520 & 0.0055 & 47.70 \\
\cmidrule(lr){2-6}
& 9:1 
& \makecell[c]{Rotation Random region masking} 
& 520 & 0.0023 &55.53 \\
\midrule[0.8pt]
\multirow{1}{*}{CLIP-Driven  Model \cite{wang2025Clip}} 
& 9:1 & \makecell[c]{Rotation Random region masking}  
& 520 & 0.0025 & - \\
\midrule[0.8pt]
\multirow{1}{*}{Unrolled ADMM \cite{10230700}} 
& 8:1 & - 
& 520 & 0.0209 & - \\
\midrule[0.8pt]
\multirow{1}{*}{NoL-MBMI  \cite{liu2024material}} 
& 8:1 &- 
& 520 &- & 31.81 \\
\bottomrule[1.5pt]
\end{tabular*}}
\end{table*}
\subsection{Comparisons with Recent State-of-the-Art Methods}
{
Table~\ref{tab10} reports a quantitative comparison {of} the proposed method {with} several recent representative  DECT material decomposition methods on the breast spectral CT dataset. For each method, the number of projection views, the {training-to-test} sample ratio, and the adopted data augmentation strategy are also listed. As can be observed, with 520 projection views, JLRM attains superior quantitative results relative to the proposed method, {as indicated by its} lower mRMSE and higher mPSNR. Nonetheless, JLRM’s performance degrades markedly as the number of projection views is reduced, whereas the proposed method maintains stable decomposition accuracy under sparse-view settings. These results indicate that DECT-DRNet can maintain high material decomposition accuracy even with severely insufficient projection data, making it more suitable for sparse-view DECT material decomposition. 

Moreover, with 520 projection views, DECT-DRNet {achieves competitive performance} compared with recent representative deep learning-based methods. Using the same 9:1 {training-to-test} split and data augmentation strategy as the CLIP-Driven Model, DECT-DRNet achieves an mRMSE of 0.0023, which is lower than {the value of} 0.0025 reported for the CLIP-Driven Model. Under an 8:1 {training-to-test} split  without data augmentation, DECT-DRNet {achieves} an mRMSE of 0.0055, substantially lower than the {value of} 0.0209 obtained by the Unrolled ADMM method. Its mPSNR reaches 47.70 dB, exceeding {the value of} 31.81 dB reported for NoL-MBMI.}

\section{Conclusion}
In this paper, we propose a dual-domain refinement network with FBP-based {Jacobian} learning for sparse-view DECT multi-material decomposition. First, we model the DECT material decomposition problem as a nonlinear {least-squares} problem with $\ell_1$ regularization. Then, we combine the proximal gradient algorithm with FBP-based {Jacobian} learning to solve this problem. In the FBP-based {Jacobian} learning module, we use FBP and {a U-Net} to approximate {the adjoint Jacobian operator in the nonlinear DECT forward model.} In the dual-domain geometric refinement module {for} solving the proximal-point subproblem, we introduce Fourier residual convolutional blocks {incorporating frequency-domain} information to achieve sparse geometric regularization. The experimental results {on the breast spectral CT dataset and the abdominal CT dataset} demonstrate that our method can effectively {suppress} noise and artifacts during material decomposition while preserving details. In future work, we will focus on several aspects. First, to further improve the {generalizability} of the proposed DECT-DRNet, we will incorporate a low-rank prior inherent in DECT images into the framework. Second, we aim to extend {our method to limited-view} DECT multi-material decomposition. Third, we will conduct a rigorous theoretical study of the contraction mapping property of the learned operator and evaluate the method using real DECT physical phantom data to investigate its practical {performance}.

\bibliographystyle{ieeetr}
\bibliography{IEEE}
\end{document}